\documentclass[10.5pt]{article}
\usepackage{memrae_report}

\newcommand{\method}{MemRAE}

\reporttitle{Short--Long Memory Cooperation for Streaming Scene Reconstruction}
\reportshorttitle{Short--Long Memory Cooperation for Streaming Scene Reconstruction}

\reportauthors{Jiayu Xiong$^{1,2}$, \quad Qi Zhang$^{1,2\dagger}$}
\reportaffil{$^1$Tongji University \qquad $^2$WorldMind Inc.}
\reportmeta{\texttt{yuinst@outlook.com}, \;\texttt{qizhang\_cs@tongji.edu.cn} \;\textperiodcentered\; $^\dagger$Corresponding author \;\textperiodcentered\; \href{https://github.com/Jiayu-Xiong/MemRAE}{\texttt{https://github.com/Jiayu-Xiong/MemRAE}}}

\begin{document}
\makereporttitle

\begin{reportabstract}
An embodied agent must render earlier scene regions after they have left its recent observations, without making every novel-view query process an expanding history. A compact working representation supports economical rendering but must continually integrate incoming views without overwriting distant structure. We present \method{}, a biologically inspired architecture built on cooperating short- and long-term scene memory. A fixed-capacity working latent remains directly renderable, while a bounded patch bank stores historical evidence alongside its source camera geometry. Pre-update rendering residuals guide patch admission, and recalled patches join recent observations at block boundaries to update the working latent. Queries decode that latent alone, moving historical retrieval and reintegration entirely out of the rendering path. Across unified 6-layer and parameter-count-matched 16-layer benchmarks spanning up to $416$-frame trajectories, \method{} establishes a favorable reconstruction--latency--memory Pareto frontier, transfers to DL3DV up to $T=256$ without DL3DV long-horizon retraining, and provides a compact working representation that also supports diffusion-based generative decoding.
\par\vspace{5pt}
\noindent\textit{Long-term memory biases short-term updates, while short-term memory filters long-term storage and recall.}
\end{reportabstract}

% ============================================================================
\section{Introduction}
\label{sec:introduction}
\begin{figure}[!t]
  \centering
  \includegraphics[width=0.72\textwidth]{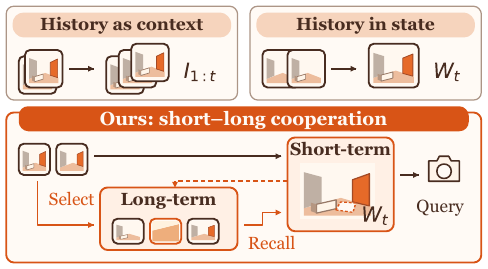}
  \caption{Streaming scene memory paradigms. Full-context methods grow linearly with the stream, while single-state recurrent models overwrite distant regions. \method{} selects patches missed by the working latent, stores them in a bounded long-term bank, and recalls them at block boundaries so queries decode the working latent alone.}
  \label{fig:teaser}
\end{figure}

% ============================================================================
An embodied agent exploring an extended 3D environment must maintain a coherent scene representation from a continuous stream of posed images~\cite{cut3r,point3r}. When it revisits earlier regions, it must synthesize novel views of places that left its field of view long ago~\cite{worldmem,vmem}. Learned scene latents enable such rendering without replaying raw frames at every query~\cite{lvsm,scenetok}. However, as trajectories span hundreds of viewpoints across multi-room layouts, neither resident memory footprint nor per-query latency can grow with exploration duration~\cite{hm3d,loopnav,streamingllm}. A streaming scene memory must therefore keep early observations usable for rendering while holding its resident state and query cost fixed (Fig.~\ref{fig:teaser}).

Streaming methods allocate historical information differently. Recurrent states and fast-weight mechanisms integrate observations into reusable representations~\cite{cut3r,ttt3r,lact,fsm}, while spatial memories and frame banks retain addressable evidence~\cite{point3r,spann3r}. Their storage and query interfaces differ, so a compact state, a selective cache, and a retrieved frame bank behave differently under extended streams~\cite{resplat,streamvggt}. Together, these approaches highlight two complementary requirements: continually updating an active scene representation and preserving evidence from earlier viewpoints~\cite{lvsm,scenetok,rae}. A memory system needs both when exploration repeatedly leaves and revisits scene regions~\cite{loopnav}. The shared challenge is retaining information that future views will require while new observations continually alter the active state~\cite{titans,bmojo}, motivating a structural separation between persistent storage and the rendering representation~\cite{teyler,winocur,kumaran2016cls}.

The hardest regime arises when querying an earlier region after long intervening exploration. A local recency window no longer contains its evidence, and a fixed working state must divide its capacity between past structure and incoming frames~\cite{parisi2019continual,streamingllm}. An external history bank preserves raw evidence, but retrieving and reading from it at every query adds latency to the rendering path~\cite{vmem,memorizing}. The gap is therefore not storage alone: it is maintaining the renderability of distant regions through a bounded active representation~\cite{kumaran2016cls} without touching the historical store at query time~\cite{coala}.

We propose \method{}, which separates continual scene integration from persistent historical storage through short--long cooperation. Inspired by biological accounts of episodic memory~\cite{marr,teyler,winocur}, \method{} maintains a directly renderable working latent alongside a bounded long-term patch bank with source geometry. Before assimilation, the current working latent predicts each incoming view to score rendering residuals and guide patch admission. At block boundaries, recalled historical patches and recent observations jointly update the working latent, which queries then decode directly without accessing or modifying the bank. Our contributions are threefold:
\begin{itemize}
  \item A biologically inspired short--long scene memory that separates continual integration from bounded historical storage while keeping a directly renderable working state.
  \item Causal observation-time admission, bounded merging and eviction, and block-boundary recall that move historical access entirely outside the query computation path.
  \item Unified 6-layer and parameter-count-matched 16-layer evaluations up to $416$ frames, controlled path switches, internal telemetry analysis, zero-retraining DL3DV transfer, and generative decoding from the working latent.
\end{itemize}

\method{} couples the two memory paths reciprocally: \textit{long-term memory biases short-term updates, while short-term memory filters long-term storage and recall.} Historical evidence fills representational gaps, while the fixed-capacity working state guides retention and recall. On the parameter-count-matched $16$-layer benchmark at $T=416$, this architecture reduces resident state by $8.2\times$ and query latency by up to $7.3\times$ with the lowest peak VRAM ($2.28\text{ GiB}$), staying within $0.4\text{ dB}$ of unbounded full-history WorldMem while outperforming bounded streaming LaCT by $+1.75\text{ dB}$ on early-view retention.

% ============================================================================
\section{Related Work}
\label{sec:related}
% ============================================================================
\paragraph{Persistent representations for streaming 3D perception.}
Persistent-state methods accumulate observations in evolving scene representations. CUT3R updates recurrent latent tokens~\cite{cut3r}, Point3R maintains explicit spatial pointers~\cite{point3r}, and Spann3R combines working and long-term spatial memory~\cite{spann3r}. ReSplat refines Gaussian primitives through rendering-error feedback~\cite{resplat}. Fast-weight approaches encode context through inference-time adaptation: LaCT uses large-chunk updates~\cite{lact}, while FSM stabilizes successive updates through elastic consolidation~\cite{fsm}. TTT3R regulates recurrent state updates to improve length generalization~\cite{ttt3r}. Across these approaches, history is compressed into latent tokens, retained as addressable features, or absorbed into adaptive weights, making long-horizon retention sensitive to capacity interference from subsequent updates.

\paragraph{Context-based synthesis and visual memory.}
Novel-view synthesis either attends to source observations during rendering or decodes a compressed scene representation. LVSM explores both direct input-view attention and a fixed latent bottleneck~\cite{lvsm}; PRoPE incorporates relative camera geometry into attention~\cite{prope}, and SceneTok compresses multiple views into scene tokens~\cite{scenetok}. Visual-memory methods retain historical observations for later conditioning: VMem retrieves views through surfel indexing~\cite{vmem}, while WorldMem retrieves historical frames and associated states for consistent world simulation~\cite{worldmem}. Explicit histories preserve fine source details, whereas latent bottlenecks provide reusable decoding interfaces. Bounding retrieved context does not bound the underlying store, making storage capacity, retrieval bandwidth, and query computation distinct costs.

\paragraph{Selective sequence memory.}
Sequence-memory methods extend recent context through adaptive compression and selective access. Titans updates neural memory using surprise~\cite{titans}, while B'MOJO combines fading state with eidetic tokens selected via an innovation test~\cite{bmojo}. EM-LLM segments streams into events and combines similarity-based with temporally contiguous retrieval~\cite{emllm}. MoM routes tokens among memory states to reduce interference~\cite{mom}, and HippoRAG supports associative retrieval through a knowledge graph~\cite{hipporag}. These methods separate compressed state from explicitly retained evidence and regulate access to each. Streaming 3D scene memory additionally requires geometric compatibility so recalled patches remain anchored to their original viewpoints across camera motion.

% ============================================================================
\section{Method}
\label{sec:method}
\begin{figure}[!t]
  \centering
  \includegraphics[width=\textwidth]{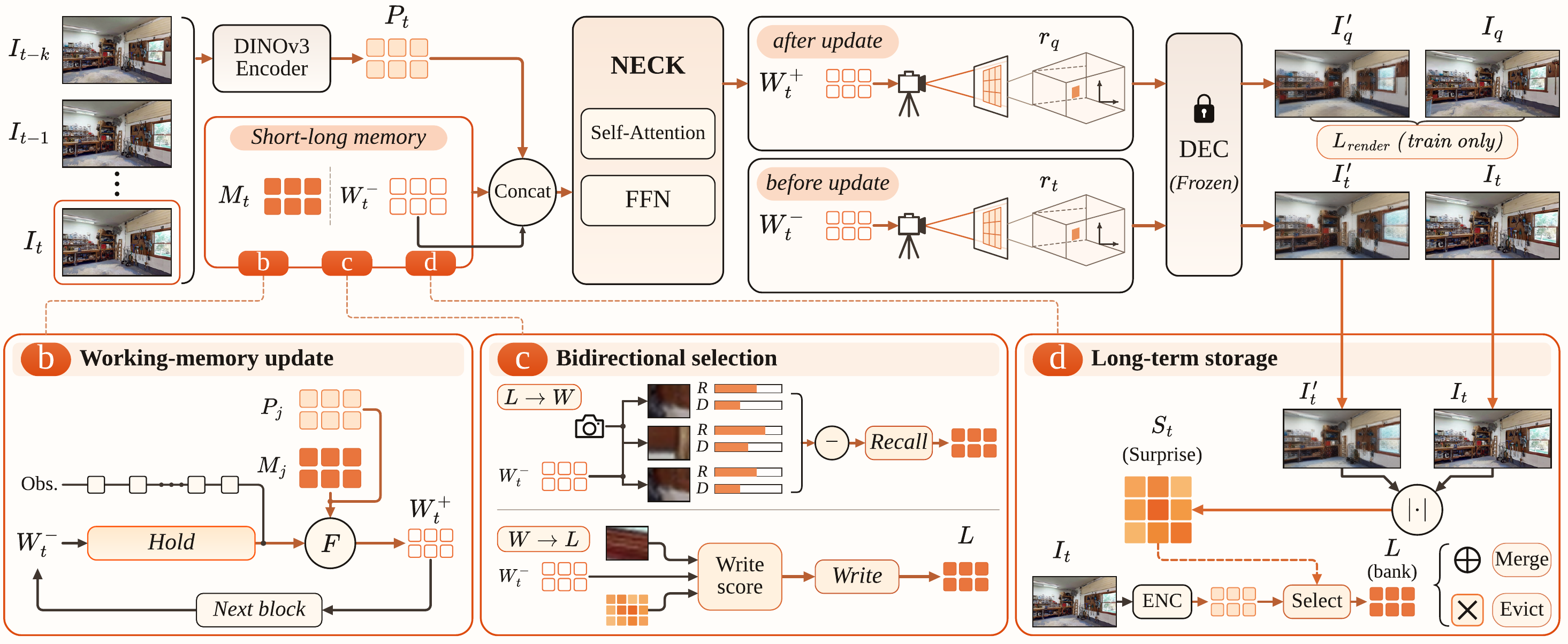}
  \caption{Short--long memory cooperation in \method{}. Pre-update rendering residuals guide patch admission into a bounded long-term bank, while merging and eviction enforce capacity. At block boundaries, recent observations and recalled historical patches jointly update the working latent $W_t$. Novel-view queries decode $W_t$ alone.}
  \label{fig:architecture}
\end{figure}

% ============================================================================
\method{} couples a renderable working latent with bounded historical patch storage (Fig.~\ref{fig:architecture}). Sec.~\ref{sec:method:prelim} defines the geometric representation and recurrent state; Sec.~\ref{sec:method:write} specifies residual-guided admission and utility refresh; Sec.~\ref{sec:method:recall} defines block-boundary recall and the query interface; and Sec.~\ref{sec:method:training} covers capacity maintenance, optimization, and causality.

\subsection{Formulation and Geometric State}
\label{sec:method:prelim}
The observation stream is $\mathcal O_t=\{(I_j,\Pi_j)\}_{j=1}^t$, where $I_j\in\mathbb R^{H\times W\times3}$ and $\Pi_j=(K_j,T_j)$ contains camera intrinsics and camera-to-world pose; query RGB never enters $\mathcal O_t$. A frozen encoder supplies patch features and a learned projection supplies neck-width appearance tokens:
\begin{equation}
  X_t=\phi(I_t)\in\mathbb R^{N\times d_e},\qquad
  Y_t=\operatorname{LN}(X_tP_X),\qquad P_X\in\mathbb R^{d_e\times d}.
\label{eq:features}
\end{equation}
Patch support $\Omega_n$ has valid subset $\Omega_n^{\rm v}$. Each stored record preserves its feature, camera pose, pixel support, source identity, source RGB (for utility refresh), and scalar utility:
\begin{equation}
\begin{aligned}
  m_{t,n}&=\mathbb I[|\Omega_n^{\rm v}|>0],\qquad \kappa_{t,n}=(t,n),\\
  p_{t,n}&=(x_{t,n},\Pi_t,\Omega_n,\Omega_n^{\rm v},\kappa_{t,n}, I_t|_{\Omega_n},v_{t,n}).
\end{aligned}
\label{eq:record}
\end{equation}

The initial camera $A=T_1$ anchors coordinates; for homogeneous pixel $\bar u$, Pl\"ucker rays encode direction and moment:
\begin{equation}
\begin{aligned}
  A^{-1}T&=(R_A,o_A),\qquad d_A(u)=\frac{R_AK^{-1}\bar u}{\|R_AK^{-1}\bar u\|_2},\\
  r_A(\Pi,u)&=[d_A(u)\,\|\,o_A\times d_A(u)].
\end{aligned}
\label{eq:rays}
\end{equation}
Under any global rigid transform $G$, $(GA)^{-1}(GT)=A^{-1}T$, leaving the ray representation invariant. Source patch coordinates retain their original spatial positions rather than being reassigned after recall. The bounded recurrent state is $\mathcal Z_t=(W_t,\mathcal M_t,\mathcal S_t,\mathcal H_t)$, initialized by a learned $W_0$ and empty buffers, satisfying
\begin{equation}
\begin{aligned}
  W_t&\in\mathbb R^{C_W\times d},\qquad |\mathcal M_t|\le C_M,\\
  |\mathcal S_t|&\le C_SN,\qquad |\mathcal H_t|\le C_H.
\end{aligned}
\label{eq:invariants}
\end{equation}
Here $\mathcal S_t$ holds the newest $C_S$ frames as patch records and $\mathcal H_t$ holds scalar residuals. A block of $B\ge C_S$ observations is processed chronologically, and queries follow complete blocks.

\subsection{Residual-Guided Historical Admission}
\label{sec:method:write}
The long-term bank preserves evidence that the working state cannot yet reconstruct. Admission uses rendering residuals computed against the pre-update state so a frame cannot suppress its own residual:
\begin{equation}
\begin{aligned}
  \widehat I^-_t&=\mathcal D(W_{t-1},r_A(\Pi_t)),\\
  s_{t,n}&=\frac{\sum_{u\in\Omega_n^{\rm v}}\|I_t(u)-\widehat I^-_t(u)\|_1}{3|\Omega_n^{\rm v}|},\qquad m_{t,n}=1.
\end{aligned}
\label{eq:surprise}
\end{equation}
High rendering error identifies patches missed by $W_{t-1}$. Let $Q_\rho$ denote the empirical quantile and $\bar w_{t-1}=C_W^{-1}\sum_j w_{t-1,j}$. Candidate writes satisfy
\begin{equation}
\begin{aligned}
  \tau_t&=Q_\rho(\mathcal H_{t-1}),\qquad
  z_{t,n}=f_w([x_{t,n}\,\|\,\bar w_{t-1}\,\|\,s_{t,n}]),\\
  \mathcal A_t&=\{n:m_{t,n}=1,\ s_{t,n}>\tau_t,\ z_{t,n}>0\},\\
  U_t&=\underset{U\subseteq\mathcal A_t,\ |U|\le K_W}{\arg\max}\sum_{n\in U}z_{t,n}.
\end{aligned}
\label{eq:write}
\end{equation}
At initialization, eligibility tests on residual and score are bypassed and up to $K_W$ valid patches are ranked by $z$. When strided admission is enabled to avoid redundant adjacent-frame writes, admission is triggered every $\Delta_{\rm adm}$-th observation; admitted records receive $v_{t,n}=s_{t,n}$ and are committed only after any recall at step $t$.

Because stored evidence can become redundant as the working latent evolves, a bounded sampled subset $\mathcal B_t\subseteq\mathcal M_{t-1}$ refreshes its utility using source-view reconstruction error:
\begin{equation}
  v_i\leftarrow\mu v_i+(1-\mu)\frac{\sum_{u\in\Omega_i^{\rm v}}\|I_i(u)-\mathcal D(W_{t-1},r_A(\Pi_i))(u)\|_1}{3|\Omega_i^{\rm v}|}.
\label{eq:value}
\end{equation}
The recent buffer and residual window are truncated to their prescribed capacities:
\begin{equation}
\begin{aligned}
  \mathcal S_t&=\operatorname{tail}_{C_SN}(\mathcal S_{t-1}\,\|\,\{p_{t,n}\}_{n=1}^N),\\
  \mathcal H_t&=\operatorname{tail}_{C_H}(\mathcal H_{t-1}\,\|\,\{s_{t,n}:m_{t,n}=1\}).
\end{aligned}
\label{eq:buffers}
\end{equation}
The admission threshold $\tau_t$ reads the pre-step window $\mathcal H_{t-1}$ before committing the updated window.

\subsection{Block-Boundary Recall and Query Decoding}
\label{sec:method:recall}
At each block boundary, recent and historical records jointly update the working latent. Let $\mathcal S_t^s$ contain the newest $C_s N$ records ($C_s<C_S$), and set $C_R=C_S N$ and $K_R=(C_S-C_s)N$. Excluding identities already in the short context yields the recall candidate set and selection:
\begin{equation}
\begin{aligned}
  \mathcal C_t&=\{p_i\in\mathcal M_{t-1}:\kappa_i\notin\kappa(\mathcal S_t^s)\},\\
  \ell_i&=f_r([x_i\,\|\,\operatorname{vec}(T_t^{-1}T_i)])-\lambda\cos(h_k(x_i),h_w(\bar w_{t-1})),\\
  \mathcal P_t&=\underset{P\subseteq\mathcal C_t,\ |P|=\min(K_R,|\mathcal C_t|)}{\arg\max}\sum_{p_i\in P}\ell_i.
\end{aligned}
\label{eq:recall}
\end{equation}
Recall scores combine relative camera geometry with a feature-alignment penalty to avoid redundant tokens. Unfilled historical slots are padded from unused recent records, newest first, forming a unique-identity context $\mathcal X_t$ of size $C_R$. For patch $p_i$, let $R_i$ denote its Pl\"ucker ray block from Eq.~\eqref{eq:rays}, $V_i$ its visibility channel, and $e_s,e_w$ learned role embeddings. Appearance and geometry enter the neck $\mathcal N$ as
\begin{equation}
\begin{aligned}
  e_i&=\operatorname{LN}(x_i P_X)+\psi([R_i;V_i])+e_s,\\
  Z_t^0&=[W_{t-1}+e_w;\{e_i\}_{p_i\in\mathcal X_t}],\\
  W_t&=\begin{cases}
    \operatorname{LN}(\mathcal N(Z_t^0;\{\Omega_i\}))_{1:C_W},& B\mid t,\\
    W_{t-1},&\text{otherwise}.
  \end{cases}
\end{aligned}
\label{eq:update}
\end{equation}
The transformer preserves original patch coordinates for spatial positional encoding. Historical evidence shapes rendering strictly through this block update rather than query-time retrieval:
\begin{equation}
\begin{aligned}
  \widehat I_{t,q}&=\mathcal D(W_t,r_A(\Pi_q)),\\
  (W_t,\mathcal M_t)_{\rm after\ query}&=(W_t,\mathcal M_t)_{\rm before\ query}.
\end{aligned}
\label{eq:query}
\end{equation}
When paired with a downstream generative renderer $\mathcal G$, the identical frozen working latent $W_t$ can alternatively condition diffusion-based synthesis $\widehat I_{t,q}^{\mathrm{gen}}=\mathcal G(W_t,r_A(\Pi_q),\epsilon)$ without modifying the streaming memory frontend.

\subsection{Bounded Maintenance, Optimization, and Guarantees}
\label{sec:method:training}
Before admitting new records, the bank reclaims $\Delta_t=\max(0,|\mathcal M_{t-1}|+|U_t|-C_M)$ slots. A low-utility pool $\mathcal L_t$ supplies candidate pairs satisfying spatial, angular, and feature compatibility:
\begin{equation}
  \|o_i-o_j\|\le\delta_o,\qquad
  \langle d_i^c,d_j^c\rangle\ge\delta_d,\qquad
  \cos(x_i,x_j)\ge\delta_x,
\label{eq:compat}
\end{equation}
where $d_i^c$ is the center-ray direction. Shortlisted pairs form $a_{ij}=[x_i\,\|\,x_j\,\|\,\operatorname{vec}(T_i^{-1}T_j)]$ with learned merge scores and fused features:
\begin{equation}
  g_{ij}=f_s(a_{ij}),\qquad
  \widetilde x_{ij}=\tfrac12(x_i+x_j)+f_m(a_{ij}).
\label{eq:merge}
\end{equation}
Positive-score pairs are greedily merged in descending order without sharing records up to $\Delta_t$, retaining one source record's geometry and the larger utility. Let $\mathcal M'$ be the resulting bank and $\Delta'_t=\max(0,|\mathcal M'|+|U_t|-C_M)$; evicting the $\Delta'_t$ lowest-utility records enforces the strict capacity bound:
\begin{equation}
  \mathcal E_t=\underset{E\subseteq\mathcal M',\ |E|=\Delta'_t}{\arg\min}\sum_{p_i\in E}v_i,\qquad
  \mathcal M_t=(\mathcal M'\setminus\mathcal E_t)\cup\{p_{t,n}:n\in U_t\}.
\label{eq:evict}
\end{equation}

During training, query reconstruction losses propagate through the recurrent block updates to optimize the memory modules. For valid target pixels $V_q$, the objective combines $\ell_1$ and perceptual loss with straight-through selection gain $\gamma(z)$:
\begin{equation}
\begin{aligned}
  \ell_q&=\frac{\sum_{u\in V_q}\|\widehat I_q(u)-I_q(u)\|_1}{3|V_q|}+\beta\ell_{\rm perc}^{V_q}(\widehat I_q,I_q),\\
  \mathcal L&=|\mathcal V|^{-1}\sum_{q\in\mathcal V}\ell_q,\qquad
  \gamma(z)=1+\sigma(z/\kappa)-\operatorname{sg}(\sigma(z/\kappa)).
\end{aligned}
\label{eq:loss}
\end{equation}
Selected records carry gain $\gamma$ during training so gradients reach write, recall, and merge scoring heads, while the merge-feature network receives ordinary reconstruction gradients.

\paragraph{Complexity and causality.}
Let $L=C_W+C_R$, neck depth $D_N$, and merge-pool size $C_L$. For fixed image resolution, memory and compute satisfy
\begin{equation}
\begin{aligned}
  M_{\rm state}&=O(C_Wd+(C_M+C_SN)(d_e+|\Omega|)+C_H),\\
  C_{\rm neck}&=O(D_N(L^2d+Ld^2)),\\
  C_{\rm recall}&=O(C_Md_eh+C_M\log K_R),\\
  C_{\rm maintain}&=O(C_L^2d_e+C_M\log C_M),
\end{aligned}
\label{eq:bounds}
\end{equation}
where $h$ is score width. Total stream processing grows linearly with trajectory length, while resident state and decoder-only query cost remain strictly bounded. For $t_b=bB$, writing $W_{t_b}=F_\theta(W_{t_{b-1}},\mathcal X_{t_b})$, recalled patches satisfy
\begin{equation}
  \mathcal P_{t_b}\subseteq\mathcal M_{t_b-1},\qquad \kappa_i=(j,n)\Longrightarrow j<t_b,
\label{eq:causal}
\end{equation}
ensuring strict prefix causality: a current patch cannot enter the historical path of its own update. At inference, binary path switches $a_s,a_l\in\{0,1\}$ isolate the carried working state and long-term bank without retraining:
\begin{equation}
\begin{aligned}
  W_b^{\rm in}&=a_s W_{t_{b-1}}+(1-a_s)W_0,\\
  \mathcal X_b^{(a_l)}&=\begin{cases}
    \mathcal X_{t_b},&a_l=1,\\
    \operatorname{tail}_{C_R}(\mathcal S_{t_b}),&a_l=0.
  \end{cases}
\end{aligned}
\label{eq:interventions}
\end{equation}
Setting $a_l=0$ fills the identical context budget $C_R$ from recent frames alone, while setting $a_s=0$ resets the carried state before each block update.

% ============================================================================
\section{Experimental Setup}
\label{sec:setup}
% ============================================================================
\paragraph{Datasets and evaluation splits.}
We evaluate across three 3D scene benchmarks. \textbf{LoopNav}~\cite{loopnav} provides $23{,}033$ training and $2{,}880$ validation trajectories (up to $128$ outward observations and $8$ return queries each, totaling $23{,}040$ queries), plus a separate $10$-scene subset for 6-layer path switches. \textbf{DL3DV}~\cite{dl3dv} contributes $50$ validation scenes with $128$ observations and $8$ held-out queries each; for zero-retraining $256$-frame transfer evaluation, we evaluate all $49$ scenes containing at least $256$ frames. \textbf{HM3D}~\cite{hm3d} supplies $20$ frozen validation scenes under two query protocols. The $6$-query/scene split (Tabs.~\ref{tab:ablation}--\ref{tab:allocation}, Fig.~\ref{fig:temporal}) uses $128$ front-camera observations with three historical and three current non-front queries per scene ($120$ queries total). The extended long-horizon evaluations (Tabs.~\ref{tab:matched16_quality}, \ref{tab:long6k}, Appendix~\ref{app:factors}) use \texttt{fixed40} ($1$ far and $1$ near non-front query per scene at $T=128$, $40$ queries total) and \texttt{long416} (fixed early and recent held-out targets tracked up to $T=416$). Target frames never enter $\mathcal O_t$. We report PSNR, SSIM~\cite{ssim}, and LPIPS~\cite{lpips}.

\paragraph{Default architecture and hyperparameters.}
All frames are resized to $480\times270$, padded to $480\times272$ for patch alignment, and cropped back before scoring. A frozen DINOv3 ViT-L/16 encoder~\cite{dinov3} ($303.1\text{M}$ parameters) yields $N=510$ patches projected from $1024$ to $d=768$ channels, and the working latent uses $C_W=1{,}024$ slots. Unless varied in an ablation, \method{} uses block size $B=16$, recent buffer $C_S=8$ frames ($C_s=4$ recent context frames), reference context budget $C_R=4{,}080$, recall budget $K_R=2{,}040$, bank capacity $C_M=16{,}320$ patches, write cap $K_W=510$, admission quantile $\rho=0.8$, and EMA $\mu=0.9$. On all extended $16$-layer HM3D long-horizon experiments, admission uses interval $\Delta_{\rm adm}=8$ frames unless swept; the zero-retraining DL3DV transfer evaluation (Tab.~\ref{tab:dl3dv_transfer}) uses $\Delta_{\rm adm}=1$. Memory heads use two-layer MLPs with $64$-dimensional hidden layers and GELU activations ($0.63\text{M}$ parameters). Under the 6-layer comparison protocol, \method{} (Ours) uses a $12$-layer neck and \method{} (Ours-L) uses a $36$-layer neck.

\paragraph{Evaluation protocols.}
We organize experiments into three clearly separated regimes:
\begin{itemize}
  \item \textbf{Unified 6-layer benchmark (Sec.~\ref{sec:res:6l}):} All streaming mechanisms are trained from scratch at $480\times272$ on a frozen DINOv3 encoder and a $6$-layer decoder to isolate memory design. Baselines span full-history or query-time readout (\textbf{Offline Prefix}, \textbf{WorldMem}~\cite{worldmem}, \textbf{LaCT}~\cite{lact}, \textbf{FSM}~\cite{fsm}, where 6-layer LaCT and FSM buffer all observed frame tokens before evaluating committed fast weights) and bounded or compressive states (\textbf{Sliding S16}, \textbf{Spann3R}~\cite{spann3r}, \textbf{TTT3R}~\cite{ttt3r}). Spann3R and TTT3R are reimplemented at the mechanism level on the shared encoder and decoder.
  \item \textbf{16-layer RAE-aligned validation (Tabs.~\ref{tab:ablation}--\ref{tab:allocation}, Figs.~\ref{fig:temporal}, \ref{fig:memory}, and~\ref{fig:rae}):} Following joint pretraining across LoopNav, DL3DV, and HM3D on the $16$-layer 3DRAE-aligned decoder ($115.80\text{M}$ decoder parameters; $203.83\text{M}$ non-DINO backbone) without per-dataset re-tuning, we verify the offline $8$-view backbone against official 3DRAE and evaluate HM3D path switches, retention, capacity sweeps, and context allocation on the $6$-query/scene split.
  \item \textbf{Extended 16-layer long-horizon comparisons (Secs.~\ref{sec:res:16l}--\ref{sec:res:long}, Secs.~\ref{sec:ana:ablations}--\ref{sec:ana:contents}, Appendix~\ref{app:factors}):} For the parameter-count-matched comparison against WorldMem and chunk-16 streaming LaCT (Sec.~\ref{sec:res:16l}), all three methods match non-DINO parameters at $\approx 204.5\text{M}$ (details in Appendix~\ref{app:factors}). They share the pretrained $115.80\text{M}$ $16$-layer decoder, initialize their memory necks from scratch, and train with full-sequence gradients to $4{,}000$ updates, where every method improves by $\le 0.4\text{ dB}$ over the prior $1{,}000$ updates. Offline and sliding bounds (\textbf{Offline-U16/U8}, \textbf{Sliding S16}, \textbf{Sliding S16 (Block)}) use this same $4{,}000$-step backbone, alongside \textbf{Offline-Full} cost scaling. Separately, pretrained-backbone \method{} checkpoints at $3{,}000$ and $6{,}000$ updates serve controlled factor ablations (Sec.~\ref{sec:ana:ablations}, Tab.~\ref{tab:app_ablation3k}) and extended scaling, dynamics, visualizations, sensitivity, and DL3DV transfer (Sec.~\ref{sec:res:long}, Secs.~\ref{sec:ana:behavior}--\ref{sec:ana:contents}, Fig.~\ref{fig:revisit_hm3d}, Tab.~\ref{tab:app_sensitivity6k}).
\end{itemize}

\paragraph{Isolated L40 profiling and generative decoding.}
Execution cost is measured on a single isolated NVIDIA L40 GPU at $\text{batch}=1$ with synchronized CUDA timing, recording \texttt{max\_memory\_allocated} as Peak VRAM and end-to-end stream ingestion including frozen DINOv3 encoding. End-to-end query latency is timed without inner synchronization, while a separate synchronized pass profiles non-overlapping FOV retrieval, neck read, and decoder stages (aggregated by per-column median across scenes). For generative decoding ($W\rightarrow\text{SD3.5 DiT}$) on LoopNav and DL3DV, we freeze the \method{} frontend emitting $W_t\in\mathbb R^{1024\times768}$ and train a pretrained SD3.5 DiT conditioned on $W_t$ and target Pl\"ucker rays via flow matching. Optimization uses AdamW ($\beta=(0.9,0.99)$, weight decay $0.01$, $1{,}000$-step warmup, learning rate $5\times10^{-5}$ on DiT blocks and $2\times10^{-4}$ on adapters), evaluated with EMA weights and $20$ Euler steps. Cross-scene $W$-swap controls hold target rays and initial noise identical while substituting $W_t$ from another validation scene in the same dataset.

% ============================================================================
\section{Main Results and Scaling Analysis}
\label{sec:results}
% ============================================================================

\subsection{Unified 6-Layer Streaming Benchmark}
\label{sec:res:6l}
{\looseness=-1 Tab.~\ref{tab:main} separates the streaming memory regimes cleanly. Full-history retrieval (WorldMem) sets the quality upper bound at high query latency and linearly growing state, whereas purely recurrent compression (TTT3R) degrades on LoopNav revisits and Sliding S16 loses geometry outside its recent window. At parameter and peak-memory footprints comparable to lightweight baselines, \method{} (Ours) outperforms all evaluated bounded and compressive methods while nearing full-history perceptual quality on DL3DV. Scaling the neck in \method{} (Ours-L) narrows the remaining gap, surpassing Offline Prefix and LaCT on LoopNav while matching WorldMem in LoopNav LPIPS and LaCT in DL3DV PSNR.}

\begin{table}[!t]
\centering
\caption{Unified 6-layer comparison of streaming mechanisms on LoopNav and DL3DV validation. Cost uses an L40 at $T=128, Q=8$; Query reports per-query latency (ms/q). Sel.\ denotes selection: \textbf{A} (all frames retained), \textbf{Q} (query-time retrieval), \textbf{W} (recency window), \textbf{S} (spatial-memory filtering), \textbf{N} (uniform recurrent update), and \textbf{R} (residual-guided write and recall). Peak, State, and Saved report base / max-configuration values. Saved counts patch-token frame-equivalents ($510\times768\times\text{FP32}$), whereas State (MiB) also includes stored RGB, masks, and geometry; Read measures dynamic memory read per query. Params exclude the shared 303.1M encoder. Bold marks the best and underline the second best ($^\dagger$non-streaming reference; $^\ddagger$36-layer neck).}
\label{tab:main}
\footnotesize
\setlength{\tabcolsep}{2.2pt}
\def\maxval#1#2{#1/{\scriptsize #2}}
\begin{tabularx}{\textwidth}{@{}Xccccccccccccc@{}}
\toprule
\tblhead
Method & Sel. & \multicolumn{3}{c}{LoopNav} & \multicolumn{2}{c}{DL3DV} & Total$\downarrow$ & Query$\downarrow$ & Peak$\downarrow$ & State$\downarrow$ & Saved & Read$\downarrow$ & Params$\downarrow$ \\
\cmidrule(lr){3-5}\cmidrule(lr){6-7}
\tblhead
& & PSNR$\uparrow$ & SSIM$\uparrow$ & LPIPS$\downarrow$ & PSNR$\uparrow$ & LPIPS$\downarrow$ & (s) & (ms/q) & (GiB) & (MiB) & (eq.) & (eq.) & (M) \\
\midrule
\multicolumn{14}{l}{\textit{Full-history retention / query-time readout}} \\
Offline Prefix$^\dagger$ & A & 14.63 & .486 & .666 & 13.18 & .7581 & 4.59 & 8.27 & \maxval{3.63}{3.63} & \maxval{192.2}{192.2} & \maxval{128}{128} & \underline{2.01} & \textbf{133.9} \\
WorldMem~\cite{worldmem} & Q & \textbf{15.20} & \textbf{.493} & \textbf{.659} & \textbf{13.46} & \textbf{.7555} & 7.43 & 99.27 & \maxval{4.22}{4.22} & \maxval{192.7}{192.7} & \maxval{128}{128} & 8.07 & 446.4 \\
LaCT~\cite{lact} & A & 14.79 & \underline{.490} & \underline{.664} & \underline{13.23} & .7611 & \underline{3.36} & 25.73 & \maxval{5.10}{5.10} & \maxval{191.3}{191.3} & \maxval{128}{128} & 108.42 & 203.8 \\
FSM~\cite{fsm} & A & 13.88 & .470 & .689 & 12.71 & .7656 & 3.87 & 30.96 & \maxval{4.61}{4.61} & \maxval{191.3}{486.0} & \maxval{128}{128} & 108.42 & 203.8 \\
\midrule
\multicolumn{14}{l}{\textit{Bounded / compressive-state}} \\
Sliding S16 & W & 14.03 & .478 & .686 & 12.85 & .7695 & \textbf{2.79} & 8.38 & \maxval{\textbf{2.05}}{\textbf{2.05}} & \maxval{24.0}{24.0} & \maxval{16}{16} & \underline{2.01} & \textbf{133.9} \\
Spann3R~\cite{spann3r} & S & 14.11 & .477 & .688 & 12.94 & .7676 & 10.02 & 9.24 & \maxval{3.13}{3.13} & \maxval{\underline{15.5}}{\underline{21.5}} & \maxval{9.8}{13.8} & 12.87 & 325.7 \\
TTT3R~\cite{ttt3r} & N & 11.06 & .444 & .734 & 12.73 & .7800 & 10.34 & 63.92 & \maxval{3.08}{3.08} & \maxval{\textbf{1.0}}{\textbf{1.0}} & N/A & \textbf{0.64} & 276.4 \\
\method{} (Ours) & R & 14.37 & .485 & .668 & 13.15 & \underline{.7559} & 4.29 & \underline{7.28} & \maxval{\underline{2.15}}{\underline{2.15}} & \maxval{71.6}{104.9} & \maxval{27.0}{40.0} & \underline{2.01} & \underline{134.5} \\
\method{} (Ours-L)$^\ddagger$ & R & \underline{14.92} & \underline{.490} & \textbf{.659} & \underline{13.23} & .7594 & 4.72 & \textbf{7.22} & \maxval{3.07}{3.07} & \maxval{26.1}{104.9} & \maxval{9.2}{40.0} & \underline{2.01} & 304.7 \\
\bottomrule
\end{tabularx}
\end{table}

\begin{figure}[!t]
  \centering
  \includegraphics[width=\textwidth]{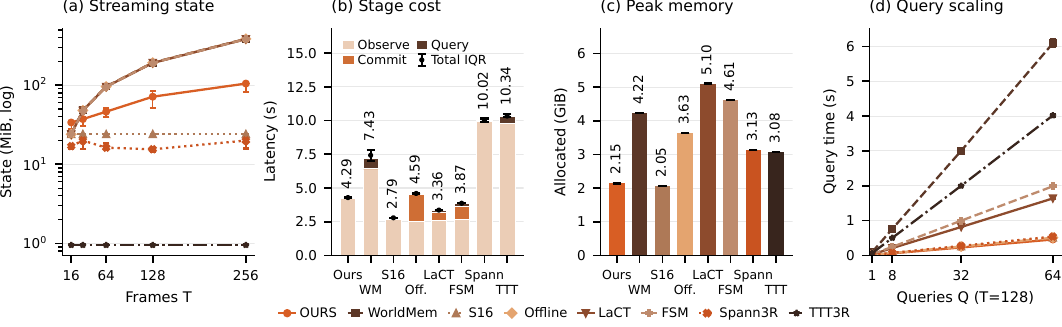}
  \caption{Cost of eight streaming mechanisms on an L40 GPU with the 6-layer decoder. (a) Resident state versus observed frames $T$. (b) Latency split into observation, commit, and query stages at $T=128, Q=8$. (c) Peak allocated VRAM. (d) Total query time versus query count $Q$ at $T=128$. Error bars show interquartile ranges over $10$ trajectories.}
  \label{fig:efficiency}
\end{figure}

Fig.~\ref{fig:efficiency} and Tabs.~\ref{tab:queryparts}--\ref{tab:dlcost} expose the stage-cost trade-offs behind these numbers. Buffering baselines (LaCT and FSM) keep ingestion fast but pay heavy commit and fast-weight read costs, while WorldMem incurs large per-view retrieval and context-reading latency. By shifting historical integration upstream into observation-time residuals and block updates, \method{} achieves nearly identical decoder latency with zero query-time retrieval, yielding $3.5\times$, $8.8\times$, and $13.6\times$ lower query latency than LaCT, TTT3R, and WorldMem at $T=128$. On short streams ($T=16$, Tab.~\ref{tab:dlcost}), residual-guided admission raises cumulative runtime above sliding-window baselines, whereas on extended or multi-query streams (Fig.~\ref{fig:efficiency}a,c,d) \method{} plateaus at bank capacity and crosses below LaCT and FSM in total runtime as $Q$ grows.

\begin{table}[!t]
\centering
\begin{minipage}[t]{0.47\textwidth}
\centering
\caption{Independent query-stage latency breakdown (total ms across $Q=8$ queries) on an L40 at $T=128, Q=8$ under the 6-layer protocol.}
\label{tab:queryparts}
\footnotesize
\setlength{\tabcolsep}{2.5pt}
\begin{tabularx}{\linewidth}{@{}Xrrrr@{}}
\toprule
\tblhead
Stage (ms) & Ours & Ours-L & LaCT & WorldMem \\
\midrule
State copy & 0.340 & 0.332 & 4.045 & 4.907 \\
Retrieval & 0.000 & 0.000 & 0.000 & 260.598 \\
Context read & 0.089 & 0.086 & 134.128 & 415.953 \\
Decoder & 56.282 & 55.551 & 55.829 & 58.671 \\
\midrule
Total & 57.910 & 57.463 & 204.138 & 750.093 \\
\bottomrule
\end{tabularx}
\end{minipage}
\hfill
\begin{minipage}[t]{0.50\textwidth}
\centering
\caption{Short-stream cost on DL3DV ($T=16$ observations, $Q=8$ queries) under the 6-layer protocol.}
\label{tab:dlcost}
\footnotesize
\setlength{\tabcolsep}{2.5pt}
\begin{tabularx}{\linewidth}{@{}Xrrrr@{}}
\toprule
\tblhead
Method & Total (s)$\downarrow$ & Obs. (ms)$\downarrow$ & Query (ms/q)$\downarrow$ & Peak (GiB)$\downarrow$ \\
\midrule
Offline Prefix$^\dagger$ & 0.449 & 20.4 & 7.26 & 2.05 \\
Sliding S16 & \textbf{0.440} & \textbf{20.0} & \textbf{7.22} & \textbf{2.05} \\
Spann3R~\cite{spann3r} & 1.181 & 69.6 & 8.48 & 3.12 \\
LaCT~\cite{lact} & \underline{0.586} & \underline{20.1} & 24.97 & 2.72 \\
FSM~\cite{fsm} & 0.717 & 20.9 & 31.27 & 3.70 \\
TTT3R~\cite{ttt3r} & 1.032 & 49.2 & 30.56 & 2.98 \\
WorldMem~\cite{worldmem} & 0.966 & 26.2 & 68.19 & 3.89 \\
\method{} (Ours) & 0.615 & 34.7 & \underline{7.49} & \underline{2.10} \\
\bottomrule
\end{tabularx}
\end{minipage}
\end{table}

\subsection{Parameter-Count-Matched 16-Layer Comparison}
\label{sec:res:16l}
\begin{figure}[!t]
  \centering
  \includegraphics[width=\textwidth]{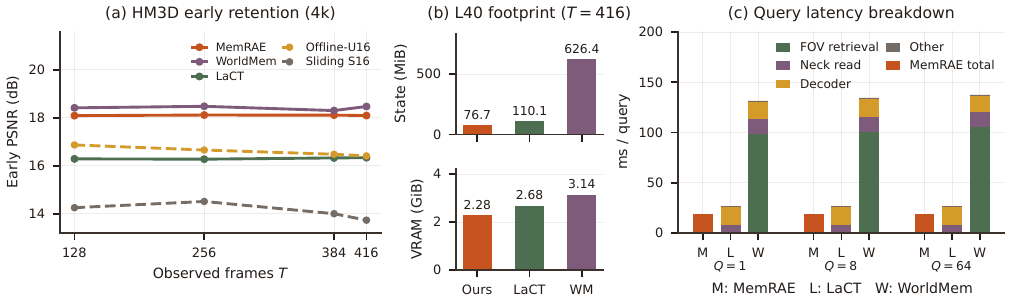}
  \caption{Parameter-count-matched HM3D comparison ($T\le 416$). (a) Early-prefix PSNR at $4{,}000$ updates for \method{}, WorldMem, chunk-16 LaCT, Offline-U16, and Sliding S16. (b) L40 resident state and peak VRAM at $T=416$. (c) Per-query latency decomposition across $Q\in\{1,8,64\}$.}
  \label{fig:matched16}
\end{figure}

Tabs.~\ref{tab:matched16_quality}--\ref{tab:matched16_cost} and Fig.~\ref{fig:matched16} evaluate \method{}, WorldMem, and chunk-16 streaming LaCT on the $16$-layer decoder up to $T=416$, characterizing the trade-off between memory footprint, query-time computation, and retention. WorldMem stores every observed frame without compression ($O(T)$ state) and runs target-conditioned FOV retrieval plus multi-frame reading at every query, forming the unbounded-retrieval upper envelope on fidelity at the cost of linear memory growth and heavy query latency. At the opposite extreme, chunk-16 LaCT folds all history into fixed-size fast weights with no explicit episodic bank. Its resident state remains bounded, with lower early-view PSNR than \method{} under this evaluation protocol, and every query still runs fast-weight neck reading ($8.4\text{ ms}$) ahead of decoding ($17.5\text{ ms}$; $26.8\text{ ms}$ total per view).

\method{} targets the bounded-state, constant-query-time frontier by consolidating observations into a query-agnostic working latent $W_t$ supported by a bounded episodic patch bank. Queries decode solely from $W_t$ with zero retrieval overhead and horizon-independent latency. Across $416$ frames under full-sequence training, \method{} maintains flat early-view retention from $T=128$ to $416$. At $T=416$, it exceeds bounded streaming LaCT, Offline-U16, and Offline-U8 by $1.75$, $1.68$, and $1.53\text{ dB}$, respectively (and Sliding S16 by $4.35\text{ dB}$), while staying within $0.4\text{ dB}$ of unbounded full-history WorldMem. Compared with WorldMem, \method{} uses $8.2\times$ smaller resident state and up to $7.3\times$ lower query latency, while achieving the lowest peak VRAM among the three streaming models (whereas Offline-Full scales steeply in both VRAM and ingestion time). On recent views, \method{} tracks Offline-U16/U8 and Sliding S16 closely, while WorldMem retrieves and attends directly to raw recent frames at query time.

\begin{table}[!t]
\centering
\caption{Parameter-count-matched HM3D quality on the 16-layer backbone (\texttt{fixed40} and \texttt{long416}) at $3{,}000$ and $4{,}000$ updates alongside offline and sliding reference bounds. Params count non-DINO weights in millions.}
\label{tab:matched16_quality}
\footnotesize
\setlength{\tabcolsep}{2.5pt}
\begin{tabularx}{\textwidth}{@{}Xrrrrrrrrrr@{}}
\toprule
\tblhead
Method & Params & \multicolumn{3}{c}{fixed40} & \multicolumn{4}{c}{long416 early PSNR} & \multicolumn{2}{c@{}}{$T{=}416$} \\
\cmidrule(lr){3-5}\cmidrule(lr){6-9}\cmidrule(lr){10-11}
\tblhead
& non-DINO & PSNR$\uparrow$ & SSIM$\uparrow$ & LPIPS$\downarrow$ & 128 & 256 & 384 & 416 & Recent$\uparrow$ & Early LPIPS$\downarrow$ \\
\midrule
\multicolumn{11}{@{}l}{\textit{Matched random-neck initialization, shared pretrained 16L decoder (3{,}000 HM3D updates)}} \\
\method{} & 204.45M & 16.33 & 0.540 & 0.493 & 17.74 & 17.83 & 17.82 & 17.80 & 14.96 & 0.462 \\
WorldMem~\cite{worldmem} & 204.46M & 17.30 & 0.558 & 0.457 & 18.10 & 18.18 & 18.06 & 18.16 & 17.03 & 0.428 \\
LaCT~\cite{lact} & 204.45M & 15.54 & 0.520 & 0.516 & 16.05 & 16.04 & 16.11 & 16.12 & 15.52 & 0.516 \\
\midrule
\multicolumn{11}{@{}l}{\textit{Matched random-neck initialization, shared pretrained 16L decoder (4{,}000 HM3D updates)}} \\
\method{} & 204.45M & 16.58 & 0.541 & 0.476 & 18.09 & 18.12 & 18.11 & 18.09 & 15.12 & 0.432 \\
WorldMem~\cite{worldmem} & 204.46M & 17.62 & 0.562 & 0.441 & 18.42 & 18.48 & 18.30 & 18.47 & 17.18 & 0.410 \\
LaCT~\cite{lact} & 204.45M & 15.79 & 0.521 & 0.514 & 16.29 & 16.28 & 16.33 & 16.34 & 15.75 & 0.513 \\
\midrule
\multicolumn{11}{@{}l}{\textit{Same-backbone offline and sliding reference bounds (4{,}000-step random-neck backbone)}} \\
Offline-U16 & 203.83M & 16.06 & 0.529 & 0.499 & 16.87 & 16.66 & 16.48 & 16.41 & 15.31 & 0.496 \\
Offline-U8 & 203.83M & 15.97 & 0.531 & 0.500 & 16.91 & 16.62 & 16.60 & 16.56 & 15.24 & 0.491 \\
Sliding S16 & 203.83M & 14.30 & 0.508 & 0.549 & 14.26 & 14.52 & 14.01 & 13.74 & 14.92 & 0.584 \\
Sliding S16 (Block) & 203.83M & 14.28 & 0.510 & 0.548 & 14.22 & 14.57 & 14.08 & 13.71 & 14.97 & 0.581 \\
\bottomrule
\end{tabularx}
\end{table}

\begin{table}[!t]
\centering
\caption{Isolated single-L40 execution cost at $T=416$ (and Offline-Full scaling across $T\in\{128,256,416\}$) on the 16-layer backbone, reported as 5-scene medians in \texttt{bfloat16}. Ingest includes frozen DINOv3 encoding. \textbf{Query} reports uninstrumented end-to-end total latency (ms) across $Q$ queries, whereas \textbf{FOV}, \textbf{Neck read}, and \textbf{Decode} report independent column medians from synchronized stage timers; Fig.~\ref{fig:matched16}c normalizes stage shares to Query$/Q$. Blank stage cells denote uninstrumented sub-timers. Bold marks the best value at $T=416$ within each $Q$.}
\label{tab:matched16_cost}
\footnotesize
\setlength{\tabcolsep}{2.5pt}
\begin{tabularx}{\textwidth}{@{}Xlrrrrrrr@{}}
\toprule
\tblhead
Method & $T/Q$ & State & Peak & Ingest & FOV & Neck read & Decode & Query \\
\tblhead
& & MiB$\downarrow$ & GiB$\downarrow$ & s$\downarrow$ & ms$\downarrow$ & ms$\downarrow$ & ms$\downarrow$ & ms$\downarrow$ \\
\midrule
\method{} & 416/1 & \textbf{76.7} & \textbf{2.28} & 10.60 & --- & --- & --- & 18.8 \\
WorldMem~\cite{worldmem} & 416/1 & 626.4 & 3.14 & 10.66 & 100.5 & 15.4 & 17.6 & 131.6 \\
LaCT~\cite{lact} & 416/1 & 110.1 & 2.68 & \textbf{9.82} & \textbf{0.0} & \textbf{8.4} & \textbf{17.5} & 26.8 \\
Offline-Full & 128/1 & 255.0 & 5.70 & 3.06 & --- & --- & --- & 17.9 \\
Offline-Full & 256/1 & 510.0 & 9.42 & 8.67 & --- & --- & --- & 18.0 \\
Offline-Full & 416/1 & 828.8 & 14.06 & 20.83 & --- & --- & --- & \textbf{17.5} \\
\midrule
\method{} & 416/8 & \textbf{76.7} & \textbf{2.28} & 10.60 & --- & --- & --- & 150.7 \\
WorldMem~\cite{worldmem} & 416/8 & 626.4 & 3.14 & 10.66 & 800.8 & 123.2 & 142.3 & 1078.1 \\
LaCT~\cite{lact} & 416/8 & 110.1 & 2.68 & \textbf{9.82} & \textbf{0.0} & \textbf{67.3} & \textbf{139.8} & 214.4 \\
Offline-Full & 128/8 & 255.0 & 5.70 & 3.06 & --- & --- & --- & 143.9 \\
Offline-Full & 256/8 & 510.0 & 9.42 & 8.67 & --- & --- & --- & 142.1 \\
Offline-Full & 416/8 & 828.8 & 14.06 & 20.83 & --- & --- & --- & \textbf{139.8} \\
\midrule
\method{} & 416/64 & \textbf{76.7} & \textbf{2.28} & 10.60 & --- & --- & --- & 1211.4 \\
WorldMem~\cite{worldmem} & 416/64 & 626.4 & 3.14 & 10.66 & 7251.0 & 999.3 & 1130.8 & 8795.3 \\
LaCT~\cite{lact} & 416/64 & 110.1 & 2.68 & \textbf{9.82} & \textbf{0.0} & \textbf{539.2} & \textbf{1119.0} & 1720.9 \\
Offline-Full & 128/64 & 255.0 & 5.70 & 3.06 & --- & --- & --- & 1170.9 \\
Offline-Full & 256/64 & 510.0 & 9.42 & 8.67 & --- & --- & --- & 1181.7 \\
Offline-Full & 416/64 & 828.8 & 14.06 & 20.83 & --- & --- & --- & \textbf{1133.1} \\
\bottomrule
\end{tabularx}
\end{table}

\subsection{Long-Horizon Retention and Zero-Retraining Transfer}
\label{sec:res:long}
\begin{figure}[!t]
  \centering
  \includegraphics[width=\textwidth]{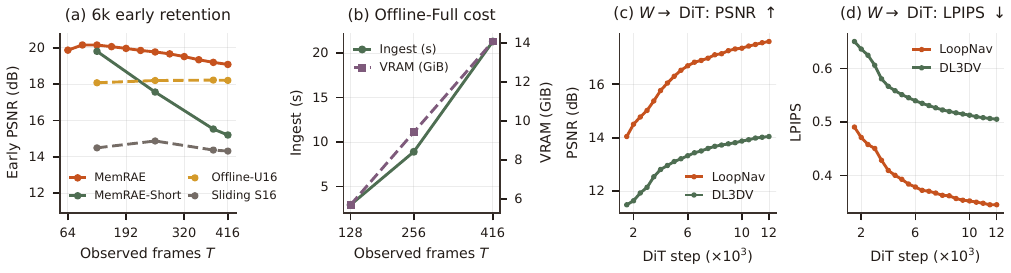}
  \caption{Long-horizon retention and generative decoding progression. (a) Dense 12-prefix early-view PSNR ($T\in[64,416]$) on HM3D versus bounds and \method{}-Short. (b) Offline-Full ingestion time and peak VRAM scaling. (c--d) Validation PSNR and LPIPS progression when training an SD3.5 DiT renderer on $W_t$ on LoopNav and DL3DV.}
  \label{fig:long_and_dit}
\end{figure}

Tabs.~\ref{tab:long6k}--\ref{tab:dl3dv_transfer} and Fig.~\ref{fig:long_and_dit}a--b report extended-horizon HM3D retention and zero-retraining DL3DV transfer. On HM3D, Sliding S16 loses distant rooms once observations exit the local window, query-time FOV Readout trails \method{} by over $1\text{ dB}$ at $T=416$ while accessing the full history, Offline-U16 requires revisiting the observed prefix, and Offline-Full incurs steeply growing ingestion time and VRAM. \method{}-Short degrades on extended streams, whereas \method{} maintains strong early-prefix retention through $T=416$.

\begin{table}[!t]
\centering
\caption{HM3D quality at $6{,}000$ updates (\texttt{fixed40} and \texttt{long416}) for \method{}, \method{}-Short, and same-backbone bounds. Bold marks the best value in each numeric column.}
\label{tab:long6k}
\footnotesize
\setlength{\tabcolsep}{2.5pt}
\begin{tabularx}{\textwidth}{@{}Xrrrrrrrr@{}}
\toprule
\tblhead
Method & fixed40 & \multicolumn{4}{c}{long416 early PSNR$\uparrow$} & Recent & SSIM$_{416}$ & LPIPS$_{416}$ \\
\cmidrule(lr){3-6}
\tblhead
& PSNR$\uparrow$ & 128 & 256 & 384 & 416 & PSNR$\uparrow$ & early$\uparrow$ & early$\downarrow$ \\
\midrule
\method{} & \textbf{18.82} & \textbf{20.16} & \textbf{19.77} & \textbf{19.20} & \textbf{19.09} & \textbf{18.93} & \textbf{0.596} & \textbf{0.374} \\
\method{}-Short & 18.66 & 19.81 & 17.56 & 15.52 & 15.20 & 16.85 & 0.480 & 0.542 \\
Offline-U16 & 17.49 & 18.08 & 18.20 & 18.22 & 18.21 & 17.98 & 0.572 & 0.389 \\
Offline-U8 & 17.64 & 18.22 & 18.39 & 18.07 & 17.92 & 18.21 & 0.567 & 0.401 \\
Sliding S16 & 14.95 & 14.49 & 14.87 & 14.37 & 14.31 & 17.43 & 0.511 & 0.533 \\
Sliding S16 (Block) & 15.03 & 14.48 & 14.90 & 14.28 & 14.27 & 17.36 & 0.515 & 0.532 \\
FOV Readout & 17.03 & 17.91 & 18.40 & 18.13 & 18.00 & 17.80 & 0.565 & 0.406 \\
\bottomrule
\end{tabularx}
\end{table}

On DL3DV ($T\in\{128,256\}$, $\Delta_{\rm adm}=1$; Tab.~\ref{tab:dl3dv_transfer}), disabling the long-term bank ($a_l=0$), resetting the carried state ($a_s=0$), or replacing selective admission and merging with a tight FIFO buffer degrades early held-out views, whereas full \method{} achieves higher early- and late-view PSNR than the evaluated ablated configurations at each stream length without additional DL3DV fine-tuning.

\begin{table}[!t]
\centering
\caption{Zero-retraining DL3DV transfer ($T\in\{128,256\}$, $\Delta_{\rm adm}=1$). Pressure counts scenes where the bank triggers patch replacement. Bold marks the best reconstruction metric per stream length.}
\label{tab:dl3dv_transfer}
\footnotesize
\setlength{\tabcolsep}{2.5pt}
\begin{tabularx}{\textwidth}{@{}Xrrrrrr@{}}
\toprule
\tblhead
Setting & Scenes & Early PSNR$\uparrow$ & Late PSNR$\uparrow$ & All PSNR$\uparrow$ & Early LPIPS$\downarrow$ & Pressure \\
\midrule
Stream 128, full & 50 & \textbf{14.78} & \textbf{14.02} & \textbf{14.40} & \textbf{0.509} & 25 \\
Stream 128, bank off ($a_l=0$) & 50 & 13.66 & 13.52 & 13.59 & 0.549 & 0 \\
Stream 128, state off ($a_s=0$) & 50 & 14.12 & 13.78 & 13.95 & 0.535 & 27 \\
Stream 128, both off & 50 & 13.14 & 13.29 & 13.21 & 0.583 & 0 \\
Stream 128, Cap 4{,}080 FIFO & 50 & 14.04 & 13.85 & 13.94 & 0.534 & 50 \\
\midrule
Stream 256, full & 49 & \textbf{13.68} & \textbf{13.68} & \textbf{13.68} & \textbf{0.558} & 49 \\
Stream 256, bank off ($a_l=0$) & 49 & 12.96 & 13.03 & 12.99 & 0.586 & 0 \\
Stream 256, Cap 4{,}080 FIFO & 49 & 13.19 & 13.37 & 13.28 & 0.580 & 49 \\
\bottomrule
\end{tabularx}
\end{table}

\subsection{Generative Decoding from Working Latents}
\label{sec:res:dit}
\begin{figure}[!t]
  \centering
  \includegraphics[width=0.95\textwidth]{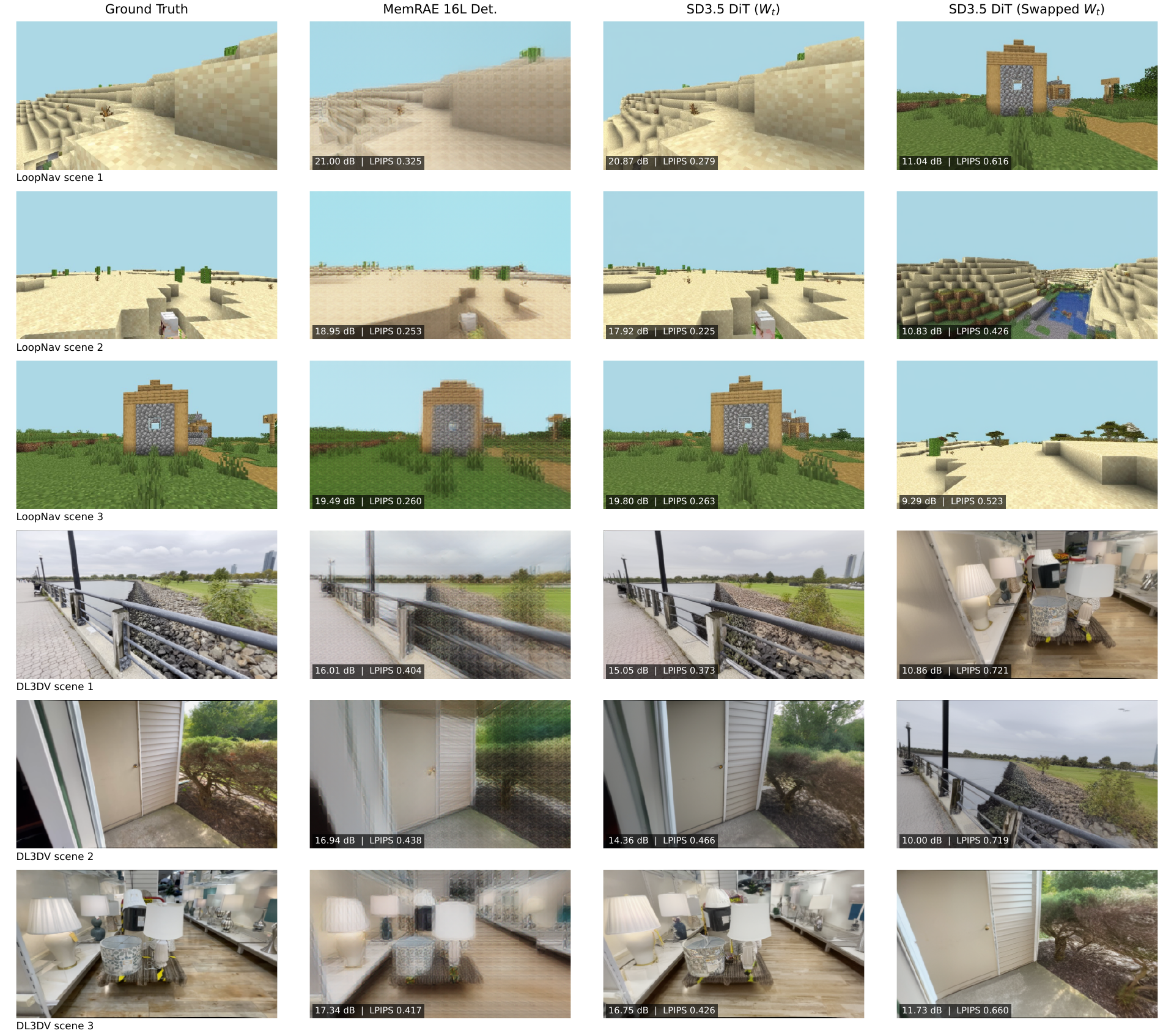}
  \caption{Qualitative $W\rightarrow\text{SD3.5 DiT}$ gallery on LoopNav and DL3DV. Columns compare Ground Truth, \method{} 16L deterministic decode, SD3.5 DiT conditioned on $W_t$, and SD3.5 DiT with a swapped cross-scene latent at identical target rays and noise.}
  \label{fig:dit_gallery}
\end{figure}

To test whether the working latent $W_t$ also supports an external generative renderer beyond deterministic decoding, we train an SD3.5 DiT head conditioned on frozen $W_t$ on LoopNav and DL3DV. Fig.~\ref{fig:long_and_dit}c--d shows steady PSNR and LPIPS improvements throughout DiT training on both datasets. Fig.~\ref{fig:dit_gallery} compares three LoopNav biomes and three DL3DV scenes. Diffusion decoding improves LPIPS in three of six examples, most clearly on LoopNav foliage and terrain textures, whereas complex DL3DV layouts can introduce generative artifacts. Swapping $W_t$ from another scene at identical target rays and initial noise lowers PSNR across all six examples. Together, these results support the use of scene-specific information from the identical frozen working latent $W_t$ across both deterministic rendering and external diffusion-based decoding without modifying the streaming memory frontend.

% ============================================================================
\FloatBarrier
\section{Analysis, Ablations, and Memory Behavior}
\label{sec:analysis}
% ============================================================================

\subsection{Short--Long Path Cooperation}
\label{sec:ana:switches}
\begin{wrapfigure}{r}{0.46\textwidth}
  \vspace{-6pt}
  \centering
  \includegraphics[width=0.45\textwidth]{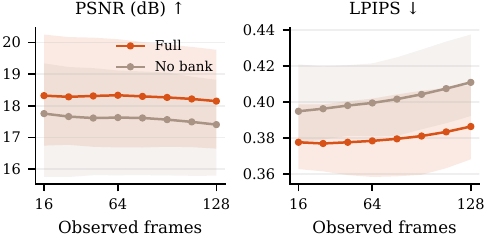}
  \caption{Fixed-target retention on HM3D ($T=16\dots 128$).}
  \label{fig:temporal}
  \vspace{-6pt}
\end{wrapfigure}
Inference-time switches $(a_s,a_l)$ from Eq.~\eqref{eq:interventions} isolate each memory path at fixed context budget $C_R$ without retraining (Tabs.~\ref{tab:sixl}--\ref{tab:ablation}). Disabling the long-term bank ($a_l=0$) reduces the model to a recent-context window plus carried working state, causing distant rooms to fade once they leave the buffer; on LoopNav and HM3D historical queries, this produces sharp PSNR drops while leaving current HM3D queries largely intact. Conversely, resetting the carried state ($a_s=0$) degrades both current and historical queries because admission, recall, and rendering all depend on the persistent working scaffold. Enabling both paths achieves the strongest overall reconstruction. Fig.~\ref{fig:temporal} confirms this cooperation along fixed-target HM3D trajectories: keeping the long-term bank active sustains higher PSNR and lower LPIPS at every prefix as new observations arrive.

\begin{table}[!t]
\centering
\begin{minipage}[t]{0.485\textwidth}
\centering
\caption{Inference-time path switches on the 6-layer comparison model (LoopNav 10-scene subset and DL3DV).}
\label{tab:sixl}
\footnotesize
\setlength{\tabcolsep}{1.8pt}
\begin{tabularx}{\linewidth}{@{}cc*{6}{>{\centering\arraybackslash}X}@{}}
\toprule
\tblhead
$a_s$ & $a_l$ & \multicolumn{3}{c}{LoopNav} & \multicolumn{3}{c@{}}{DL3DV} \\
\cmidrule(lr){3-5}\cmidrule(lr){6-8}
\tblhead
 & & PSNR$\uparrow$ & SSIM$\uparrow$ & LPIPS$\downarrow$ & PSNR$\uparrow$ & SSIM$\uparrow$ & LPIPS$\downarrow$ \\
\midrule
$\checkmark$ & $\checkmark$ & \textbf{13.554} & \textbf{.450} & .703 & \textbf{13.145} & \textbf{.364} & \textbf{.756} \\
$\checkmark$ & $\times$ & 12.397 & .444 & \textbf{.701} & 12.835 & .363 & .759 \\
$\times$ & $\checkmark$ & 13.429 & .444 & .722 & 12.230 & .351 & .759 \\
$\times$ & $\times$ & 11.291 & .414 & .759 & 11.994 & .348 & .765 \\
\bottomrule
\end{tabularx}
\end{minipage}
\hfill
\begin{minipage}[t]{0.485\textwidth}
\centering
\caption{Inference-time path switches on the 16-layer backbone (HM3D 6-query/scene split: current vs.\ historical queries).}
\label{tab:ablation}
\footnotesize
\setlength{\tabcolsep}{1.8pt}
\begin{tabularx}{\linewidth}{@{}cc*{6}{>{\centering\arraybackslash}X}@{}}
\toprule
\tblhead
$a_s$ & $a_l$ & \multicolumn{3}{c}{Current} & \multicolumn{3}{c@{}}{History} \\
\cmidrule(lr){3-5}\cmidrule(lr){6-8}
\tblhead
 & & PSNR$\uparrow$ & SSIM$\uparrow$ & LPIPS$\downarrow$ & PSNR$\uparrow$ & SSIM$\uparrow$ & LPIPS$\downarrow$ \\
\midrule
$\checkmark$ & $\checkmark$ & \textbf{18.286} & .564 & \textbf{.423} & \textbf{18.146} & \textbf{.572} & \textbf{.386} \\
$\checkmark$ & $\times$ & 18.088 & \textbf{.568} & .426 & 17.405 & .562 & .411 \\
$\times$ & $\checkmark$ & 15.442 & .507 & .519 & 15.558 & .504 & .480 \\
$\times$ & $\times$ & 15.180 & .508 & .522 & 14.355 & .469 & .559 \\
\bottomrule
\end{tabularx}
\end{minipage}
\end{table}

\subsection{Capacity Pressure, Policies, and Context Allocation}
\label{sec:ana:ablations}
\begin{table}[!t]
\centering
\begin{minipage}[t]{0.53\textwidth}
\centering
\caption{HM3D bank-capacity sweep ($T=128$, 16-layer backbone, 6-query/scene split). Pressure counts scenes triggering replacement.}
\label{tab:capacity}
\footnotesize
\setlength{\tabcolsep}{2.0pt}
\begin{tabularx}{\linewidth}{@{}*{6}{>{\raggedleft\arraybackslash}X}@{}}
\toprule
\tblhead
Patches & PSNR$\uparrow$ & SSIM$\uparrow$ & LPIPS$\downarrow$ & State (MiB)$\downarrow$ & Pressure \\
\midrule
4080 & 18.2004 & \textbf{.5684} & .4051 & \textbf{59.45} & 20/20 \\
8160 & \textbf{18.2229} & .5681 & \textbf{.4047} & 79.92 & 18/20 \\
16320 & 18.2162 & .5681 & .4048 & 98.42 & 2/20 \\
32640 & 18.2136 & .5681 & .4049 & 98.42 & 0/20 \\
\bottomrule
\end{tabularx}
\end{minipage}
\hfill
\begin{minipage}[t]{0.44\textwidth}
\centering
\caption{HM3D context allocation ($C_S=8$ frames within fixed budget $C_R$, 6-query/scene split).}
\label{tab:allocation}
\footnotesize
\setlength{\tabcolsep}{2.0pt}
\begin{tabularx}{\linewidth}{@{}*{4}{>{\raggedleft\arraybackslash}X}@{}}
\toprule
\tblhead
Recent & Recalled & History PSNR$\uparrow$ & Current PSNR$\uparrow$ \\
\midrule
8 & 0 & 17.405 & 18.088 \\
7 & 1 & 17.661 & 18.219 \\
6 & 2 & 17.897 & \textbf{18.311} \\
4 & 4 & 18.146 & 18.286 \\
1 & 7 & \textbf{18.297} & 18.241 \\
\bottomrule
\end{tabularx}
\end{minipage}
\end{table}

\begin{figure}[!tb]
  \centering
  \includegraphics[width=\textwidth]{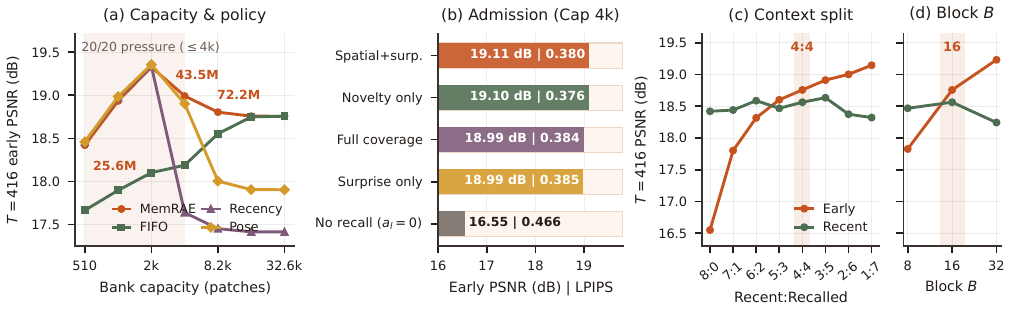}
  \caption{Controlled ablations on HM3D ($T=416$). (a) Capacity scaling under replacement pressure versus FIFO, Recency, and Pose policies, with shaded full-pressure regime and annotated resident state. (b) Admission-rule comparison at capacity $4{,}080$ alongside the no-recall control (PSNR and LPIPS). (c) Short-to-long context split ($8{:}0$ to $1{:}7$). (d) Update block size $B\in\{8,16,32\}$.}
  \label{fig:ablations16l}
\end{figure}
\looseness=-1
We next examine how \method{} behaves under capacity pressure, policy variations, and context-budget splits, pairing the $T=128$ HM3D sweeps (Tabs.~\ref{tab:capacity}--\ref{tab:allocation}) with extended $T=416$ ablations (Fig.~\ref{fig:ablations16l}; full factor matrices appear in Appendix~\ref{app:factors}). Sweeping bank capacity across nearly two orders of magnitude triggers replacement in every scene at tight budgets while preserving strong early-view quality: utility-based eviction outperforms FIFO under tight capacity ($C_M\le 8{,}160$), whereas learned recall outperforms recency and pose-only recall once capacity exceeds the recall budget ($C_M\ge 4{,}080$; Fig.~\ref{fig:ablations16l}a). Comparing the evaluated admission rules at capacity $4{,}080$ alongside the no-recall control (Fig.~\ref{fig:ablations16l}b) shows that spatially stratified surprise and novelty-aware admission lead among the tested rules in both fidelity and perceptual quality. Finally, sweeping the recent-to-recalled split within $C_R$ and varying update block size $B$ (Fig.~\ref{fig:ablations16l}c--d, Tab.~\ref{tab:allocation}) reveals a clean trade-off: larger recalled shares monotonically boost early-view reconstruction, whereas balanced splits and a moderate block size optimize joint current and historical quality.

\subsection{Observation-Step and Admission-Interval Dynamics}
\label{sec:ana:behavior}
\begin{figure}[!tb]
  \centering
  \includegraphics[width=\textwidth]{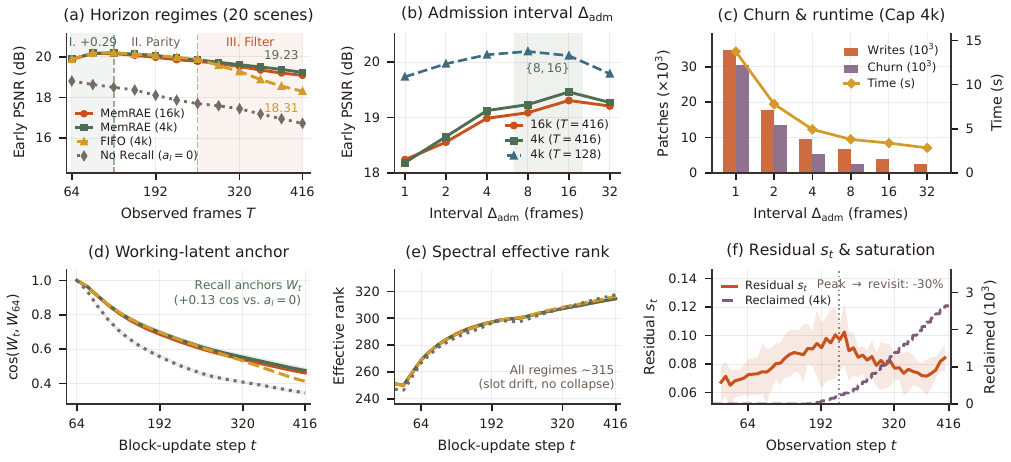}
  \caption{Observation-step and admission-interval dynamics across $20$ HM3D validation scenes. (a) Three horizon regimes separating multi-view consolidation ($T\le 128$), pre-saturation parity, and post-saturation pressure filtering. (b--c) Long-term admission interval $\Delta_{\rm adm}\in\{1,2,4,8,16,32\}$ quality versus replacement churn and memory-update stream runtime (excluding frozen DINOv3 encoding). (d--e) Anchor cosine retention $\cos(W_t,W_{64})$ and spectral effective rank across block updates. (f) Scene-conditioned pre-update residual damping and bank replacement.}
  \label{fig:horizon_dynamics}
\end{figure}
Across all $20$ HM3D validation trajectories, three distinct regimes emerge as observation count $T$ grows (Fig.~\ref{fig:horizon_dynamics}a). In \textit{Regime~I} ($T\le 128$, multi-view consolidation), fixed early-view fidelity rises rather than decays because overlapping viewpoints refine room geometry in $W_t$, whereas disabling recall ($a_l=0$) already begins losing early structure. In \textit{Regime~II} ($128<T\le 256$, pre-saturation parity), tight-capacity \method{} ($C_M=4{,}080$) and FIFO track closely until every scene reaches bank capacity near the trajectory midpoint. In \textit{Regime~III} ($T>256$, post-saturation filtering), FIFO drops steeply as age-based eviction erases early rooms, whereas \method{} ($C_M=4{,}080$) holds decay nearly flat and slightly outperforms the larger-capacity bank ($C_M=16{,}320$) at $40\%$ lower state footprint.

Sweeping the long-term admission interval $\Delta_{\rm adm}\in\{1,2,4,8,16,32\}$ reveals a pronounced inverted-U trade-off (Fig.~\ref{fig:horizon_dynamics}b--c). Admitting patches on every frame ($\Delta_{\rm adm}=1$) floods the bank with redundant adjacent-frustum tokens, triggering heavy eviction churn, up to $3.6\times$ slower memory-update stream runtime (excluding frozen DINOv3 encoding), and over $1\text{ dB}$ lower $T=416$ retention at $C_M=4{,}080$ relative to $\Delta_{\rm adm}\in\{8,16\}$ (with $\Delta_{\rm adm}=2$ showing intermediate degradation). Conversely, spacing admissions to $\Delta_{\rm adm}\in\{8,16\}$ acts as a geometric baseline filter: $\Delta_{\rm adm}=8$ peaks on medium-horizon ($T=128$) retention by supplying two baseline snapshots per $B=16$ update block, while $\Delta_{\rm adm}=16$ maximizes $T=416$ retention by nearly eliminating premature landmark eviction before $\Delta_{\rm adm}=32>B$ introduces viewpoint blind spots during rapid turns. Because admission is closed-loop (gated by pre-update residuals against $W_t$ and ray-coverage novelty against $\mathcal M_t$), saturating at $C_M=4{,}080$ around $t\approx 192$ also lowers late-stream writes relative to the unsaturated $16\text{k}$ bank ($6.7\text{k}$ vs.\ $9.7\text{k}$ cumulative writes at $\Delta_{\rm adm}=8$, with $2.6\text{k}$ merged or evicted to keep occupancy at $4{,}080$).

\begin{figure}[!tb]
  \centering
  \includegraphics[width=\textwidth]{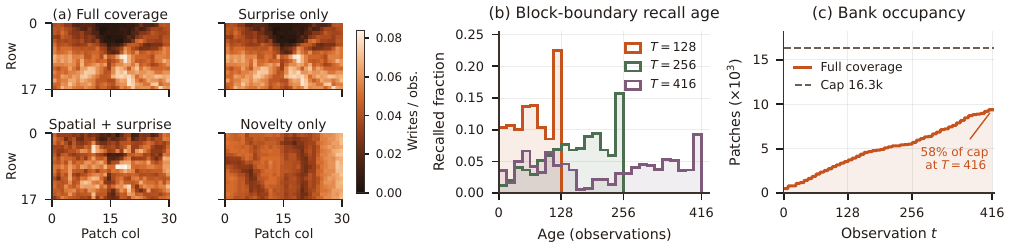}
  \caption{Patch-grid admission and recall telemetry on HM3D ($T=416$). (a) Spatial patch-write density across the $17\times 30$ grid under four admission rules. (b) Source-frame ages of patches recalled at block-boundary updates at $T\in\{128,256,416\}$. (c) Example-trajectory bank occupancy, reaching $58\%$ of capacity at $T=416$.}
  \label{fig:behavior}
\end{figure}

Tracking $W_t$ across block updates clarifies how single-state models forget (Fig.~\ref{fig:horizon_dynamics}d--e): spectral effective rank rises smoothly to $\sim 315$ across all regimes including No Recall, ruling out aggregate rank collapse and indicating that forgetting accompanies directional slot drift ($\cos(W_t,W_{64})$ dropping sharply without recall). Meanwhile, scene-conditioned pre-update residuals peak near trajectory turnaround and drop by $30\%$ as return paths revisit familiar geometry (Fig.~\ref{fig:horizon_dynamics}f). Complementing these multi-scene dynamics, spatial write-density maps and recall-age histograms (Fig.~\ref{fig:behavior}) show distinct spatial admission patterns across the $17\times30$ grid under surprise-only versus spatially stratified admission while block-boundary updates continue to recall $300+$ frame-old observations and example-trajectory bank occupancy reaches $58\%$ of capacity at $T=416$.

\subsection{Visualizing Working Memory and the Long-Term Bank}
\label{sec:ana:contents}
\begin{figure}[!tb]
  \centering
  \includegraphics[width=\textwidth]{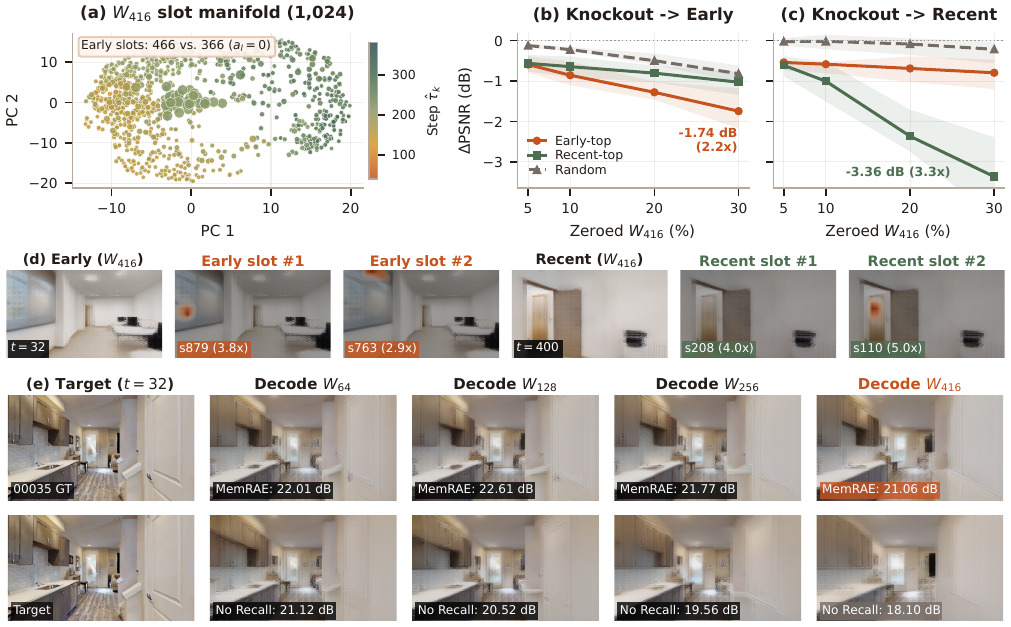}
  \caption{Inside working memory $W_{416}$. (a) 2D PCA of the $1{,}024$ latent slots colored by decoder attention-weighted step $\hat\tau_k$ and sized by total attention mass. (b--c) Causal slot-knockout asymmetry across $10$ HM3D scenes on early and recent novel views. (d) Spatial receptive fields of specialized slots when decoding early ($t=32$) and recent ($t=400$) views from the identical $W_{416}$. (e) Fixed early-target rendering progression from $W_{64}\to W_{416}$ comparing \method{} against No Recall ($a_l=0$).}
  \label{fig:w_contents}
\end{figure}
To inspect what $W_t\in\mathbb R^{1024\times768}$ encodes internally, we extract exact last-four-layer decoder cross-attention from query rays to latent slots at $T=416$ (Fig.~\ref{fig:w_contents}). Projecting all $1{,}024$ tokens via PCA colored by attention-weighted timestamp $\hat\tau_k$ reveals a crescent-shaped distribution in the 2D PCA projection spanning early-room tokens, central shared-hub tokens, and recent-room tokens, with block-boundary recall preserving substantially more early-preferring slots than No Recall ($466$ vs.\ $366$ on loop scene \texttt{00067}, and $447$ vs.\ $236$ on outward scene \texttt{00035}; Fig.~\ref{fig:w_contents}a). While shared hub tokens induce high overall linear correlation across viewpoints, rank-order selectivity diverges sharply ($\text{Spearman}=0.24$, top-$20\%$ $\text{Jaccard}=0.24$). Causally zeroing top early-attended versus recent-attended slots across $10$ scenes induces strong selective degradation (Fig.~\ref{fig:w_contents}b--c). Knocking out $30\%$ of early-specialized slots hurts early views $2.2\times$ more than recent views, whereas knocking out recent-specialized slots hurts recent views $3.3\times$ more than early views and far exceeds random knockouts. Individual slots exhibit spatially localized decoder attention around structures such as window blinds, wall corners, doorways, and office fixtures when decoding from the identical $W_{416}$ (Fig.~\ref{fig:w_contents}d). Across $W_{64}\to W_{416}$, the full model retains crisp cabinetry, sinks, and door hardware where No Recall degrades into blur (Fig.~\ref{fig:w_contents}e).

\begin{figure}[!tb]
  \centering
  \includegraphics[width=\textwidth]{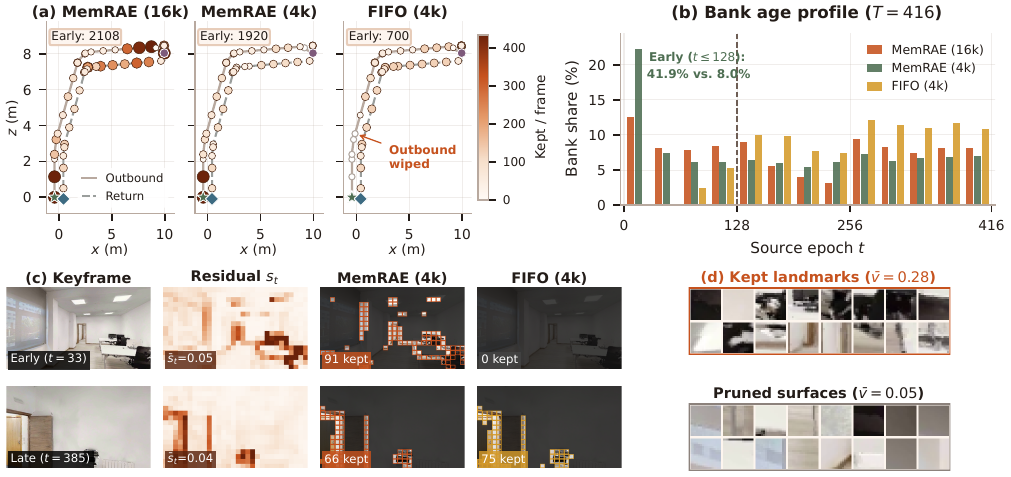}
  \caption{Inside the long-term patch bank $\mathcal M_{416}$. (a) Two-lane outbound/return bird's-eye camera trajectory comparing surviving patch counts per frame across \method{} ($C_M=16{,}320$), \method{} ($C_M=4{,}080$), and FIFO ($C_M=4{,}080$). (b) Source-epoch distribution of surviving bank patches across all $20$ HM3D scenes. (c) Keyframe pre-update residuals and surviving $16\times16$ patch overlays on early and late rooms. (d) Actual $16\times16$ RGB tiles from the early room contrasting high-utility surviving landmarks against low-utility pruned surfaces.}
  \label{fig:bank_contents}
\end{figure}
The bank visualizations show how utility-based retention and FIFO preserve different parts of the trajectory (Fig.~\ref{fig:bank_contents}). Separating outbound and return lanes in bird's-eye view (Fig.~\ref{fig:bank_contents}a) reveals a directional pitfall of FIFO: walking back along a visited corridor reverses camera orientation by $180^\circ$, so evicting outbound frames wipes out the forward-facing frustums required to render early targets. Across all $20$ scenes, \method{} under tight capacity ($C_M=4{,}080$) preserves $41.9\%$ of its bank in the early prefix ($t\le 128$, versus $8.0\%$ for FIFO) while maintaining balanced coverage across subsequent epochs (Fig.~\ref{fig:bank_contents}b). Image-space overlays and raw $16\times16$ RGB patch mosaics from $\mathcal M_{416}$ (Fig.~\ref{fig:bank_contents}c--d) show that top-utility surviving records concentrate on structural landmarks such as monitors, window mullions, desk corners, and door frames ($5.4\times$ higher mean utility than the lowest-utility pruned tiles), whereas homogeneous walls, shades, floors, and ceilings are pruned away.

\FloatBarrier
\subsection{Qualitative Grounding and Backbone Verification}
\label{sec:ana:qual}
\begin{figure}[!tb]
  \centering
  \includegraphics[width=0.97\textwidth]{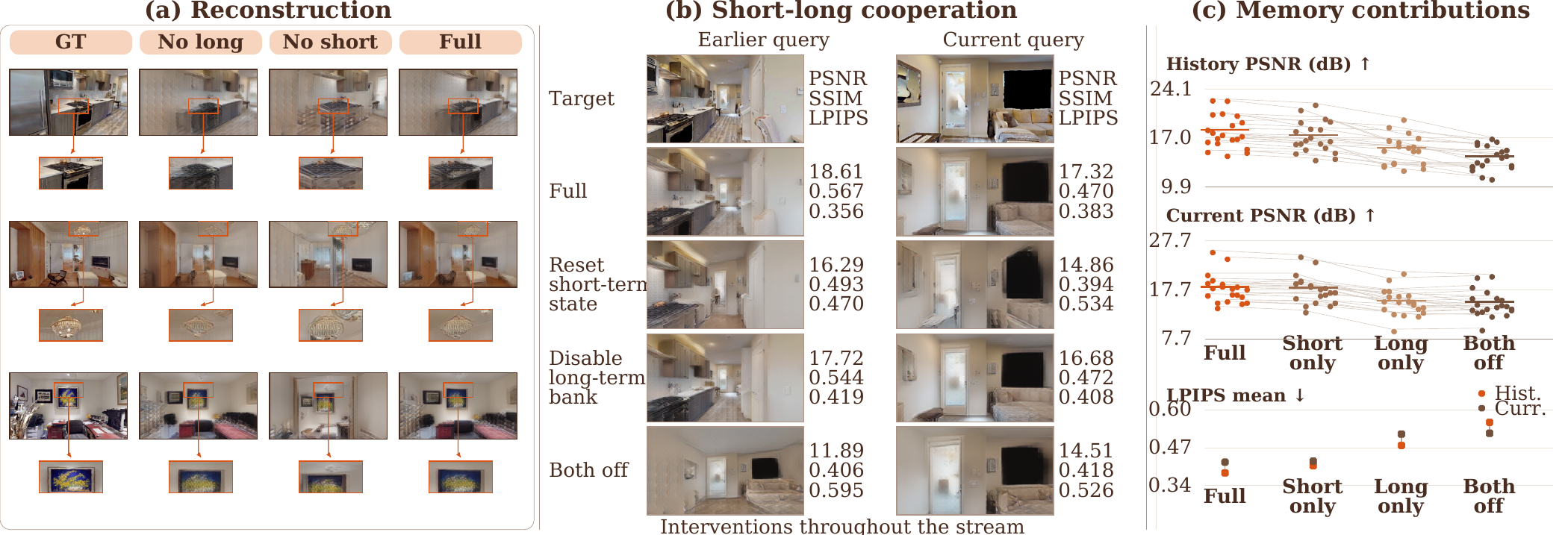}
  \caption{Qualitative and per-scene path-switch analysis on HM3D ($6$-query/scene split). (a) Fine-detail preservation under full cooperation. (b) Asymmetric failures on historical versus current views. (c) Per-scene effects of the memory-path switches across $20$ HM3D scenes.}
  \label{fig:memory}
\end{figure}
\begin{figure}[!tb]
  \centering
  \includegraphics[width=0.97\textwidth]{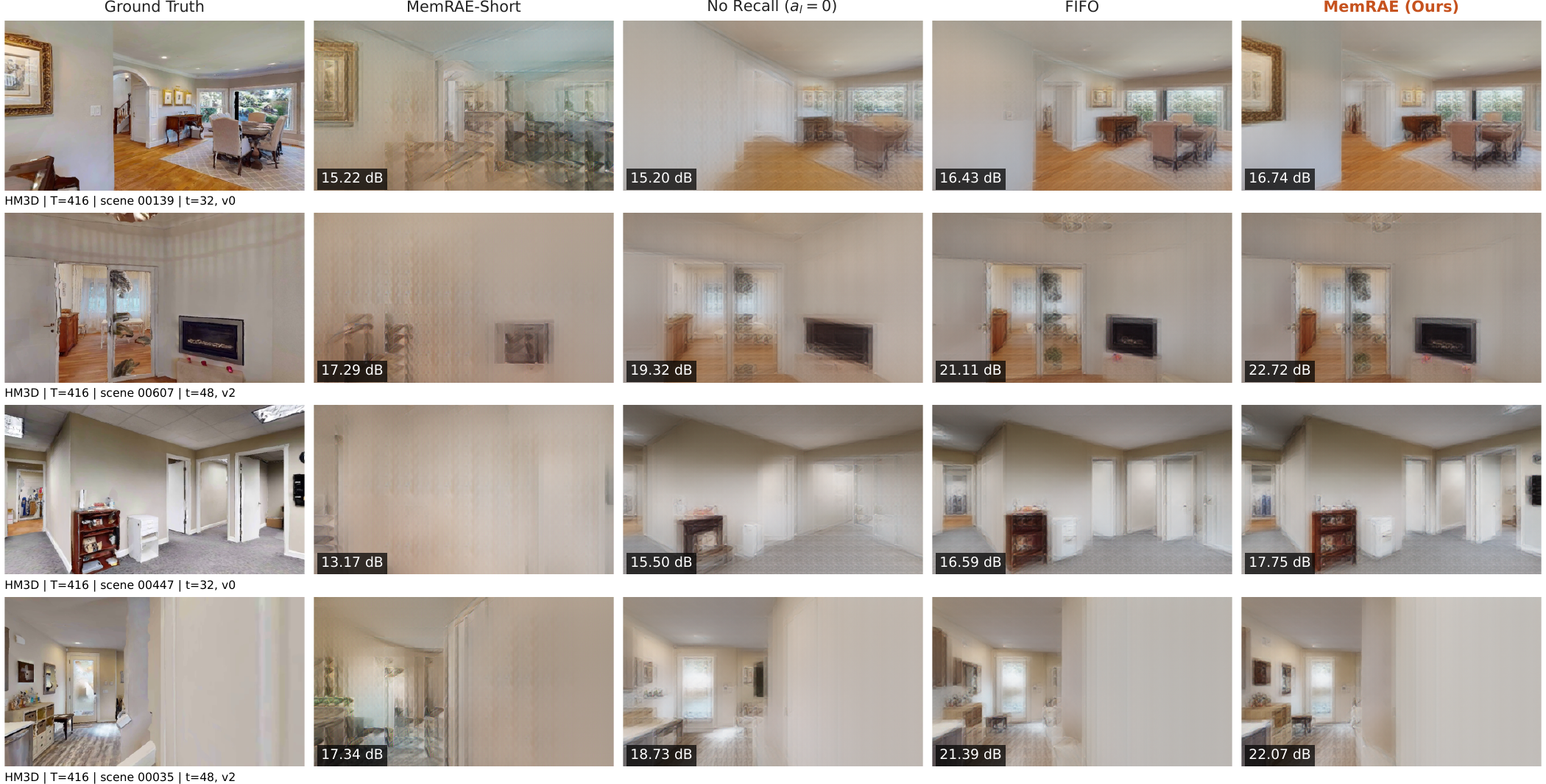}
  \caption{Qualitative long-horizon revisit comparisons across four distinct HM3D scenes at $T=416$. Columns compare Ground Truth, \method{}-Short, No Recall ($a_l=0$), FIFO, and \method{} (Ours), with per-view PSNR annotated.}
  \label{fig:revisit_hm3d}
\end{figure}
\begin{figure}[!tb]
  \centering
  \includegraphics[width=0.97\textwidth]{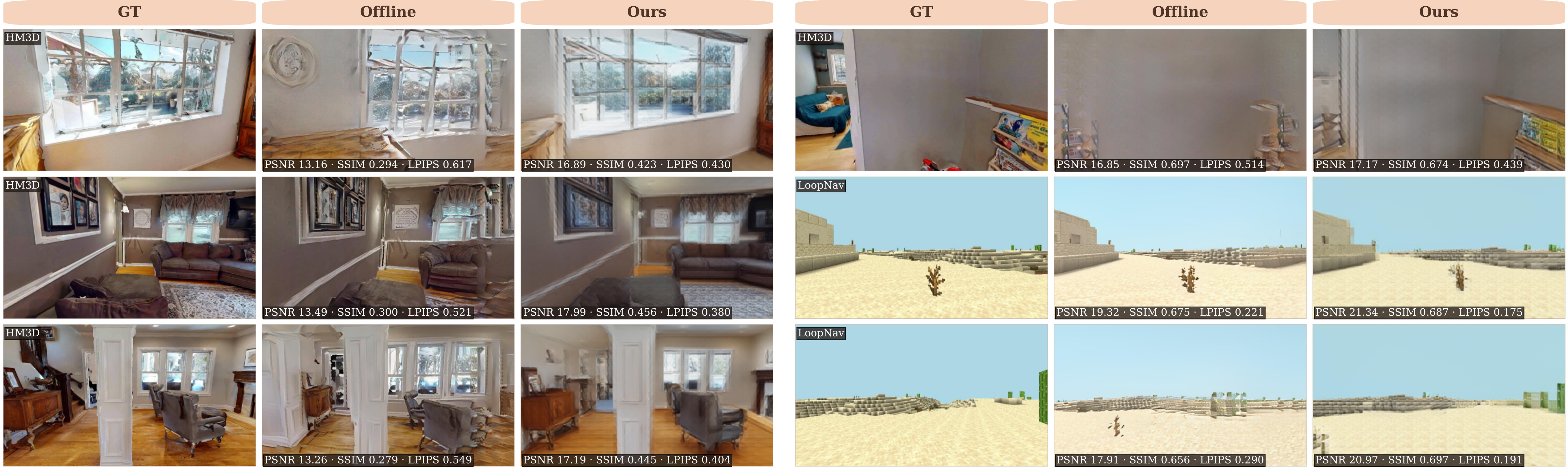}
  \caption{Offline 8-view backbone verification without streaming memory: official 3DRAE (Offline) versus our adapted 16-layer backbone (Ours) on HM3D and LoopNav novel views.}
  \label{fig:rae}
\end{figure}

Fig.~\ref{fig:memory} illustrates path-switch behavior on HM3D: disabling either memory path blurs fine furniture, lighting, and wall details, while turning both off collapses room layout, with consistent metric shifts across all $20$ validation scenes. On extended $T=416$ revisits across four distinct HM3D rooms (Fig.~\ref{fig:revisit_hm3d}), \method{}-Short and No Recall ($a_l=0$) lose distant wall art, fireplaces, and shelving once early viewpoints exit the local window, and at matched tight capacity ($C_M=4{,}080$), FIFO blurs fine structure and erases wall art whereas \method{} recovers clearer room geometry solely from $W_t$. Finally, Fig.~\ref{fig:rae} verifies the adapted $16$-layer backbone before enabling streaming memory: under identical $8$-view offline inputs, our backbone preserves cleaner window mullions, doorways, and room layouts than official 3DRAE on both HM3D and LoopNav.

% ============================================================================
\section{Conclusion and Limitations}
\label{sec:conclusion}
% ============================================================================
Long-term scene memory can preserve historical evidence without placing historical retrieval on the query rendering path. \method{} pairs a directly renderable working latent with a bounded geometric patch bank, using pre-update residuals to guide admission and block-boundary recall to refresh the working state. This establishes a transferable architectural principle for streaming 3D reconstruction: separate persistent storage from the active state consumed by the renderer.

\paragraph{Limitations and outlook.}
Our formulation assumes posed, static 3D streams where rendering residuals indicate missing structural memory; under moving objects or odometry drift, dynamic discrepancies can enter the admission signal, motivating uncertainty-aware weighting and dynamic gating. In addition, shifting historical integration to observation time favors multi-query navigation workloads, whereas single-shot streams with very few queries can incur lower total latency with lazy buffering. Finally, selective merging remains lossy under tight capacity on arbitrarily long trajectories, and scaling generative $W\rightarrow\text{SD3.5 DiT}$ decoding to complex multi-room indoor layouts remains an exciting direction for future work.

\FloatBarrier
{\small
\setlength{\bibsep}{1.8pt plus 0.3pt}
\bibliographystyle{abbrvnat}
\bibliography{refs}
}

\clearpage
\appendix
\section{Extended Factor Matrices and Implementation Details}
\label{app:factors}
Tab.~\ref{tab:app_ablation3k} provides the full $3{,}000$-step factor matrix underlying Fig.~\ref{fig:ablations16l}, and Tab.~\ref{tab:app_sensitivity6k} reports $6{,}000$-step operating-point sensitivity on HM3D ($T\le 416$). For the parameter-count-matched $16$-layer comparison (Sec.~\ref{sec:res:16l}, seed $303$), all three models share the pretrained $115.80\text{M}$ decoder and match non-decoder parameters ($\approx 88.65\text{M}$). Specifically, \method{} uses a $12$-layer neck plus $0.63\text{M}$ heads ($204.45\text{M}$), and WorldMem uses a $3$-layer retrieval-attention neck ($d=768$, $16$ heads, MLP ratio $2.556$, retrieved length $8$; $204.46\text{M}$). Chunk-16 streaming LaCT uses a $6$-layer fast-weight neck ($d=768$, $16$ heads, intermediate multiplier $2.717$, $1$ fast head, $5$ Muon inner updates; $204.45\text{M}$), keeping resident state bounded at $T=416$ unlike the full-stream $6$-layer LaCT reference in Tab.~\ref{tab:main}.

\subsection{Capacity-Pressure and Policy Regimes}
\label{app:sub:pressure}
At $3{,}000$ updates (Tab.~\ref{tab:app_ablation3k}), the capacity sweep shows that reducing bank capacity can preserve or improve early-view quality over the tested range. At the same bank capacity ($C_M=4{,}080$), spatially stratified surprise and novelty-driven retention achieve higher $T=416$ early-view PSNR than surprise-only admission ($19.11$ and $19.10$ vs.\ $18.99\text{ dB}$). Context-budget allocation and update block size then mediate the remaining early-versus-recent trade-off: larger recalled shares and coarser block updates preserve distant structure longer, whereas balanced splits keep the working latent responsive to incoming viewpoints.

\begin{table}[!htbp]
\centering
\caption{Extended 3{,}000-step factor matrix for \method{} on HM3D ($T=416$). Groups cover path switches, the capacity ladder, four admission policies (cap 4{,}080) and a no-recall reference ($a_l=0$), context allocation, and update block size. Switch columns apply only to the switch group. Bold marks the best quality value within a group.}
\label{tab:app_ablation3k}
\scriptsize
\setlength{\tabcolsep}{2.5pt}
\begin{tabularx}{\textwidth}{@{}lXccrrrrrrrr@{}}
\toprule
\tblhead
Factor & Setting & $a_s$ & $a_l$ & PSNR$_{128}$$\uparrow$ & PSNR$_{416}$$\uparrow$ & Recent$\uparrow$ & SSIM$_{416}$$\uparrow$ & LPIPS$_{416}$$\downarrow$ & State & Press. & $n$ \\
\tblhead
& & & & early & early & PSNR & early & early & MiB & scenes & scenes \\
\midrule
switch & both off & $0$ & $0$ & 14.38 & 14.38 & 17.27 & 0.521 & 0.535 & 23.0 & 0 & 20 \\
switch & bank only & $0$ & $1$ & 17.58 & 16.52 & 17.77 & 0.528 & 0.470 & 71.3 & 0 & 20 \\
switch & state only & $1$ & $0$ & 18.48 & 16.55 & 18.42 & 0.556 & 0.466 & 23.0 & 0 & 20 \\
switch & both on & $1$ & $1$ & \textbf{19.92} & \textbf{18.76} & \textbf{18.56} & \textbf{0.589} & \textbf{0.388} & 72.2 & 1 & 20 \\
\midrule
capacity & Cap 510 & --- & --- & 19.40 & 18.42 & 18.45 & 0.587 & 0.397 & 25.6 & 20 & 20 \\
capacity & Cap 1{,}020 & --- & --- & 19.63 & 18.94 & 18.53 & 0.592 & 0.385 & 28.2 & 20 & 20 \\
capacity & Cap 2{,}040 & --- & --- & 19.81 & \textbf{19.33} & 18.45 & \textbf{0.598} & \textbf{0.371} & 33.3 & 20 & 20 \\
capacity & Cap 4{,}080 & --- & --- & \textbf{19.95} & 18.99 & \textbf{18.65} & 0.591 & 0.384 & 43.5 & 20 & 20 \\
capacity & Cap 8{,}160 & --- & --- & 19.92 & 18.80 & 18.55 & 0.589 & 0.387 & 63.2 & 14 & 20 \\
capacity & Cap 16{,}320 & --- & --- & 19.92 & 18.76 & 18.56 & 0.589 & 0.388 & 72.2 & 1 & 20 \\
capacity & Cap 32{,}640 & --- & --- & 19.92 & 18.76 & 18.56 & 0.590 & 0.388 & 72.5 & 0 & 20 \\
\midrule
policy & Full coverage & --- & --- & 19.95 & 18.99 & \textbf{18.65} & 0.591 & 0.384 & 43.5 & 20 & 20 \\
policy & Surprise only & --- & --- & \textbf{19.97} & 18.99 & 18.61 & 0.591 & 0.385 & 43.5 & 20 & 20 \\
policy & Spatial+surprise & --- & --- & 19.95 & \textbf{19.11} & 18.57 & 0.593 & 0.380 & 43.5 & 20 & 20 \\
policy & Novelty only & --- & --- & 19.84 & 19.10 & 18.21 & \textbf{0.596} & \textbf{0.376} & 43.5 & 20 & 20 \\
policy & No recall & --- & --- & 18.48 & 16.55 & 18.42 & 0.556 & 0.466 & 23.0 & 0 & 20 \\
\midrule
allocation & 8:0 & --- & --- & 18.48 & 16.55 & 18.42 & 0.556 & 0.466 & 72.3 & 1 & 20 \\
allocation & 7:1 & --- & --- & 19.40 & 17.80 & 18.44 & 0.573 & 0.418 & 72.1 & 1 & 20 \\
allocation & 6:2 & --- & --- & 19.68 & 18.32 & \textbf{18.59} & 0.582 & 0.403 & 72.2 & 1 & 20 \\
allocation & 4:4 & --- & --- & 19.92 & 18.76 & 18.56 & 0.589 & 0.388 & 72.2 & 1 & 20 \\
allocation & 2:6 & --- & --- & 20.00 & 19.00 & 18.38 & 0.592 & 0.380 & 71.9 & 1 & 20 \\
allocation & 1:7 & --- & --- & \textbf{20.02} & \textbf{19.15} & 18.32 & \textbf{0.594} & \textbf{0.376} & 71.7 & 1 & 20 \\
\midrule
block & B=8 & --- & --- & 19.75 & 17.83 & 18.47 & 0.573 & 0.420 & 71.7 & 1 & 20 \\
block & B=16 & --- & --- & \textbf{19.92} & 18.76 & \textbf{18.56} & 0.589 & 0.388 & 72.2 & 1 & 20 \\
block & B=32 & --- & --- & 19.74 & \textbf{19.23} & 18.24 & \textbf{0.596} & \textbf{0.370} & 72.7 & 1 & 20 \\
\bottomrule
\end{tabularx}
\end{table}

\subsection{Operating-Point Sensitivity at 6,000 Steps}
\label{app:sub:sensitivity}
At $6{,}000$ updates (Tab.~\ref{tab:app_sensitivity6k}), all inference-time factors vary smoothly around the default operating point ($\rho=0.8$, $B=16$, $4{:}4$ allocation, $C_M=16{,}320$) without retraining. Raising the admission quantile prunes marginal writes to shrink resident state while slightly improving long-horizon fidelity, whereas lowering it admits redundant patches that inflate memory without improving early recall. Similarly, increasing the update block size slows working-state turnover to favor distant-room retention at a modest cost to recent-view sharpness, showing a smooth trade-off between memory footprint, early recall, and recency responsiveness over the tested settings.

\begin{table}[!htbp]
\centering
\caption{Operating-point sensitivity for \method{} on HM3D ($T\le 416$). Quantile, block size, and recent:recalled allocation are inference-time factors evaluated without retraining. Bold marks the best value per group.}
\label{tab:app_sensitivity6k}
\footnotesize
\setlength{\tabcolsep}{3pt}
\begin{tabularx}{\textwidth}{@{}lXrrrr@{}}
\toprule
\tblhead
Factor & Level & PSNR$_{128}$$\uparrow$ & PSNR$_{416}$$\uparrow$ & Recent$\uparrow$ & State \\
\tblhead
& & early & early & PSNR & MiB$\downarrow$ \\
\midrule
quantile & Full coverage ($\rho{=}0.80$) & 20.16 & 19.09 & \textbf{18.93} & 71.6 \\
quantile & Quantile 0.50 & 20.10 & 18.93 & 18.90 & 99.1 \\
quantile & Quantile 0.70 & 20.10 & 19.02 & 18.86 & 82.4 \\
quantile & Quantile 0.90 & \textbf{20.18} & \textbf{19.22} & 18.81 & \textbf{59.2} \\
\midrule
block & Block 8 & 20.04 & 18.16 & 18.85 & \textbf{71.3} \\
block & Block 16 & \textbf{20.16} & 19.09 & \textbf{18.93} & 71.6 \\
block & Block 32 & 19.95 & \textbf{19.61} & 18.57 & 72.9 \\
\midrule
allocation & 8:0 & 18.50 & 16.73 & 18.55 & 72.2 \\
allocation & 7:1 & 19.56 & 18.03 & 18.72 & 71.8 \\
allocation & 6:2 & 19.92 & 18.60 & 18.83 & 71.8 \\
allocation & 4:4 & 20.16 & 19.09 & \textbf{18.93} & \textbf{71.6} \\
allocation & 2:6 & \textbf{20.25} & 19.32 & 18.58 & 71.7 \\
allocation & 1:7 & 20.24 & \textbf{19.49} & 18.53 & \textbf{71.6} \\
\bottomrule
\end{tabularx}
\end{table}

\end{document}